\documentclass{article}

\usepackage{microtype}
\usepackage{graphicx}
\usepackage{subfigure}
\usepackage{booktabs} % for professional tables
\usepackage{pifont}
\usepackage{tablefootnote}
\usepackage{enumitem}

\usepackage{basic_commands}
\usepackage{hyperref}

\usepackage[accepted]{icml2024}

\usepackage{amsmath}
\usepackage{amssymb}
\usepackage{mathtools}
\usepackage{amsthm}

\usepackage[capitalize,noabbrev]{cleveref}

\theoremstyle{plain}

\theoremstyle{definition}

\theoremstyle{remark}

\DeclarePairedDelimiterX{\kldivx}[2]{(}{)}{%
  #1\;\delimsize\|\;#2%
}
\newcommand{\kldiv}{D_\mathrm{KL}\kldivx}

\usepackage[textsize=tiny]{todonotes}

\newcommand{\cutsectionup}{\vspace{-6pt}}
\newcommand{\cutsectiondown}{\vspace{-4pt}}
\newcommand{\cutsubsectionup}{\vspace{-5pt}}
\newcommand{\cutsubsectiondown}{\vspace{-4pt}}

\expandafter\def\expandafter\normalsize\expandafter{%
    \normalsize
    \setlength\abovedisplayskip{5pt}
    \setlength\belowdisplayskip{5pt}
    \setlength\abovedisplayshortskip{0pt}
    \setlength\belowdisplayshortskip{0pt}
}

\icmltitlerunning{}

\begin{document}

\twocolumn[
% \icmltitle{Inverse Decision Transformer}
\icmltitle{Unsupervised Zero-Shot Reinforcement Learning \\ via Functional Reward Encodings}
%%SL.12.23: We should brainstorm titles more. A few things to consider:
% 1. The current title kind of has a lot of technical stuff, might be perceived as only of interest to the narrow range of people that care about the intersection of: generalization, goals, multi-task learning, offline RL.
% 2. Can we have some pithier name that really captures the key idea? E.g., Learning all Tasks by Conditioning on Random Reward Functions (maybe a little too literal...)
% 3. If we can work in something about transformers or something that would be nice... somehow craft a title that communicates that this is an approach for large scale scalable unsupervised learning of all tasks, foundation model for RL, etc...



\icmlsetsymbol{equal}{*}

\begin{icmlauthorlist}
\icmlauthor{Kevin Frans}{b}
\icmlauthor{Seohong Park}{b}
\icmlauthor{Pieter Abbeel}{b}
\icmlauthor{Sergey Levine}{b}
\end{icmlauthorlist}

\centering{$^{1}$ University of California, Berkeley \\ \texttt{kvfrans@berkeley.edu}}

\icmlaffiliation{b}{University of California, Berkeley}

\icmlcorrespondingauthor{Kevin Frans}{kvfrans@berkeley.edu}

\icmlkeywords{Machine Learning, ICML}

\vskip 0.3in
]




\begin{abstract}
\vspace{-2pt}
Can we pre-train a generalist agent from a large amount of unlabeled offline trajectories such that it can be immediately adapted to any new downstream tasks in a zero-shot manner?
In this work, we present a \emph{functional} reward encoding (FRE) as a general, scalable solution to this \emph{zero-shot RL} problem.
Our main idea is to learn functional representations of any arbitrary tasks by encoding their state-reward samples using a transformer-based variational auto-encoder.
This functional encoding not only enables the pre-training of an agent from a wide diversity of general unsupervised reward functions, but also provides a way to solve any new downstream tasks in a zero-shot manner, given a small number of reward-annotated samples.
We empirically show that FRE agents trained on diverse random unsupervised reward functions can generalize to solve novel tasks in a range of simulated robotic benchmarks, often outperforming previous zero-shot RL and offline RL methods. Code for this project is provided at: \href{https://github.com/kvfrans/fre}{github.com/kvfrans/fre}.

\end{abstract}

\vspace{-22pt}
\section{Introduction}
\cutsectiondown

A useful agent is one that can accomplish many objectives in a domain. Household robots are more beneficial the more chores they can complete; self-driving cars the more places they can reach. Building upon this premise, we draw inspiration from the recent success of \emph{unsupervised learning} in language~\citep{brown2020language} and vision~\citep{kirillov2023segment}, which has shown that a single generalist model trained on Internet-scale data can immediately solve a wide array of tasks without further training or fine-tuning. Motivated by these successes, we study an analogous way to train a generalist agent from unlabeled offline data such that it can immediately solve new user-specified tasks in a without training. This has been referred to as the \emph{zero-shot reinforcement learning (RL)} problem~\citep{touati2022does}. From this data, the hard challenge is how to discover, without labels, a task representation that is robust to downstream objectives -- in essence, bypassing the need for a human to specify well-shaped reward functions before training.

In this work, we aim to provide a simple, scalable approach to the zero-shot RL problem. Our key insight is to directly learn a latent representation that can represent any arbitrary reward \emph{functions} based on their \emph{samples} of state-reward pairs. We refer to this idea as \textbf{Functional Reward Encoding} (\textbf{FRE}). This is in contrast to previous works in zero-shot RL or multi-task RL that employ domain-specific task representations~\citep{barreto2017successor, li2020generalized} or highly restrictive linear reward structures~\citep{borsa2018universal, touati2021learning, touati2022does}. By directly encoding reward \emph{functions} into a latent space, we can pre-train a multi-task agent with a host of unsupervised reward functions of arbitrary diversity, and quickly identify the representations corresponding to new test tasks given a small number of reward-annotated samples.

Training an FRE requries utilizing a prior distribution over reward functions. When no information about downstream tasks is available, we must define a prior that broadly spans possible objectives in a domain-agnostic manner. In our experiments, we show that a mixture of \textit{random unsupervised reward functions}, such as goal-reaching and random MLP rewards, are a reasonable choice for the reward prior. We optimize an FRE-conditioned policy towards all rewards within this space. In this way, approximate solutions to many downstream tasks have \textit{already been learned}, and the zero-shot RL problem reduces to simply locating the FRE encoding for the task, which the learned encoder accomplishes.

% Training an FRE requires utilizing a prior distribution over reward functions. When no information about downstream tasks is available, we must define a prior that broadly spans possible objectives without utilizing domain knowledge. In our experiments, we show that a mixture of \textit{random functions} are a reasonable choice for the reward prior. We optimize an FRE-conditioned policy towards all rewards within this space. In this way, approximate solutions to downstream tasks have \textit{already been learned}, and the problem reduces to simply locating the FRE embedding for the task, which the learned encoder accomplishes.

% An FRE-conditioned policy is trained to optimize all objectives within this space. When a downstream task is provided, its reward function is encoded via the FRE and the policy is executed without further learning. 

\begin{figure}
    \centering
    \includegraphics[width=0.49\textwidth]{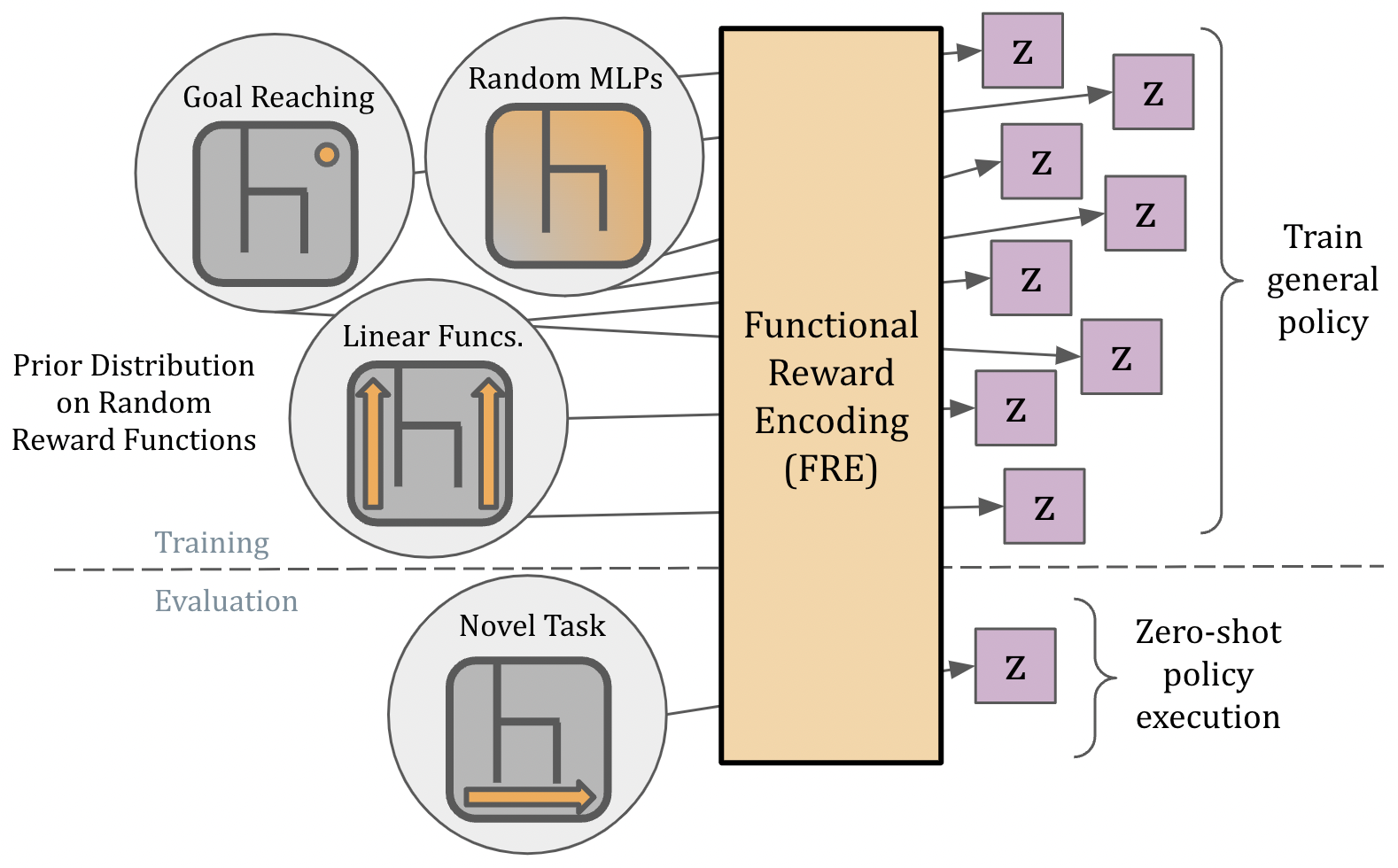}

    \caption{\textbf{FRE discovers latent representations over random unsupervised reward functions. At evaluation, user-given downstream objectives can be encoded into the latent space to enable zero-shot policy execution.} FRE utilizes simple building blocks and is a data-scalable way to learn general capabilities from unlabeled offline trajectory data.}
    \label{fig:unify}
\end{figure}

Thus, our framework presents a simple yet scalable method for training zero-shot RL agents in an unsupervised manner, as shown in Figure \ref{fig:unify}. The main idea is to (1) train an FRE network over random unsupervised reward functions, then (2) optimize a generalist FRE-conditioned policy towards maximizing said rewards, after which (3) novel tasks can be solved by simply encoding samples of their reward functions, such that the FRE agent can immediately act without further training.

\vspace{5cm}

We verify the efficacy of our method through experiments on standard offline RL domains. We demonstrate that without any finetuning, FRE policies can 
%%SL.1.29: be careful with "zero shot" (it's not... but there is nuance)
solve tasks involving locomotion of an eight-DoF robot through a maze or manipulation of a robotic arm in a kitchen scene, and can learn diverse, useful policies from the unsupervised ExORL dataset consisting of non-expert trajectories. FRE-based agents match or outperform state-of-the-art offline RL methods. Prior methods display competitive performance on either goal-reaching or structured rewards, but not both; FRE is the first method to consistently solve tasks across the board.

\cutsectionup
\section{Related Work}
\cutsectiondown

%%SL.12.23: use sentence case for paragraph headings
\textbf{Task-conditioned reinforcement learning.} Our work builds off the field of
%%SL.12.23: contextual?
multi-task RL \cite{caruana1997multitask}, where a single generalist policy is conditioned on a task description. Universal Value Functions \cite{schaul2015universal} provide a general framework for value functions conditioned on task descriptions, which are sometimes referred to as `metadata' \cite{sodhani2021multi} or contextual MDPs \cite{hallak2015contextual}.
% Task-conditioning often considers only a given family of tasks, and has taken the form of policy sketches \cite{andreas2017modular} and language information \cite{shridhar2023perceiver, silva2021lancon}, among others \cite{reed2022generalist}.
Previous multi-task RL methods typically assume a \emph{parameterized} family of tasks, specifying tasks by manually designed task parameters~\citep{barreto2017successor}, symbolic annotations~\citep{andreas2017modular}, or language embeddings~\citep{silva2021lancon, shridhar2023perceiver}.
% In this work, we bypass the need for domain-specific task-conditioning entirely. By instead learning a latent encoding over unsupervised rewards, we can generalize to objectives without assuming a specific task family.
Unlike these works, we bypass the need for domain-specific task-conditioning entirely. By instead learning a \emph{functional} latent encoding over unsupervised rewards, we can express and generalize to any arbitrary reward functions, without assuming a parameterized task family.

\textbf{Zero-shot reinforcement learning with successor features.}
% Our framework considers the zero-shot RL paradigm,
% whose goal is to pre-train a multi-task policy such that it can directly be adapted to unseen tasks without any further training steps.
Previous works have presented successor feature (SF)-based solutions to the zero-shot RL problem~\citep{dayan1993improving, barreto2017successor, borsa2018universal, chen2023self}, where they learn universal value functions based on a linear combination of pre-defined or learned state features.
%%SP: I feel Gamma models per se are not zero-shot RL methods.
% Gamma-models \cite{janner2020gamma} show that generative models over future states can also be utilized.
Approaches such as the forward-backward method \cite{touati2021learning, touati2022does} extend the SF framework by avoiding the need to train state features separately.
% However, the assumption of linearity in SF methods makes them tricky to train and limited in scope.
% Similarly to SF methods, our work utilizes the functional form of reward functions for task-conditioning. In contrast, we bypass the assumption of linearity, keeping the universality of SF methods while simplifying the learning setup. In contrast to SF methods which require on the order of thousands of state-reward pairs during evaluation, our method requires only 32 to 256.
% However, the assumption of linearity in these SF-based methods may significantly limit the types of tasks that the multi-task policy can learn.
However, these SF-based methods share a limitation in that they can only learn value functions in the linear span of state features.
In contrast, we do not make any assumptions about the task structure, allowing the policy to learn any reward functions based on our functional reward embedding. We show that this not only makes our policy \emph{universal}, but also leads to better empirical performance than these SF methods in our experiments.

 % Goal-conditioned RL \cite{kaelbling1993learning} focuses on a subset of reward functions that involve reaching states in the lowest amount of time. A number of specialized algorithms utilize this structure to perform better value learning \cite{nair2018visual, eysenbach2022contrastive, wang2023optimal} or utilize hindsight for efficient trajectory relabelling \cite{andrychowicz2017hindsight, park2023hiql}.
% Our method considers goal-conditioned RL as a subset of the full zero-shot RL problem. We examine goal-conditioned tasks as both an unsupervised reward family and as a downstream task, and further generalize to other reward families. \red{cite GCRL that finetune from a GCRL policy?}

\textbf{Goal-conditioned reinforcement learning.} Goal-conditioned RL~\citep{kaelbling1993learning} provides another way to train a multi-task policy,
whose aim is to learn to reach any goal states in the lowest amount of time.
There have been proposed a variety of methods for online~\citep{andrychowicz2017hindsight, levy2017learning, nair2018visual, savinov2018semi, fang2018dher, durugkar2021adversarial, agarwal2023f}
and offline~\cite{chebotar2021actionable, yang2022rethinking, eysenbach2022contrastive, li2022hierarchical, wang2023optimal, park2023hiql} goal-conditioned RL.
In this work, we consider goal-conditioned RL as a subset of the full zero-shot RL problem:
we train our policy with a more general set of unsupervised reward families that include goal-conditioned tasks.
As a result, our policy learns much more diverse behaviors than goal-reaching,
which is crucial for solving general reward functions at test time, as we will show in our experiments. 

\textbf{Unsupervised skill learning.}
Our method is related to previous online and offline unsupervised skill learning methods, as we also train a multi-task policy from offline data in an unsupervised manner. Online unsupervised skill discovery methods train skill policies by maximizing various intrinsic rewards~\citep{eysenbach2018diversity, sharma2019dynamics, strouse2021learning, laskin2022cic, park2023metra}. Offline skill learning methods train multi-task policies via behavioral cloning on trajectory chunks~\citep{ajay2020opal, pertsch2021accelerating} or offline RL with random reward functions~\citep{hu2023unsupervised}. These methods, however, either assume high-quality demonstrations or do not provide an efficient mechanism to adapt to tasks at test time. In contrast, our approach trains diverse policies that are optimal for a wide array of reward functions, while jointly learning a functional reward encoding that enables zero-shot test-time adaptation.
% with no assumptions on dataset quality.

\textbf{Offline Meta-RL.}
Finally, our problem setting is conceptually related to offline meta RL,
whose goal is to learn to solve tasks efficiently at test time by training an agent on diverse tasks or environments.
Previous works in meta-RL and offline meta-RL have proposed diverse techniques,
such as permutation-invariant task encoders,
to encode tasks into a latent space~\citep{duan2016rl, rakelly2019efficient, li2020focal, li2020multi, dorfman2021offline, pong2022offline, yuan2022robust},
similarly to FRE.
% and they also often employ permutation-invariant task encoders as in FRE.
However, these offline meta-RL methods typically assume a set of tasks and task-specific datasets;
on the contrary, we focus on the \emph{unsupervised} zero-shot RL setting, where we are only given a single unlabeled dataset, without assuming datasets compartmentalized by tasks or any form of reward supervision.

% This can also be interpreted as an extreme case of meta-RL, where we do not have access to meta-training tasks.
% %% SP: Not sure if I like the following sentence
% Hence, we aim to \emph{both define and encode} a set of unsupervised or unsupervised reward functions for zero-shot task adaptation at test time.

%%SL.12.23: can we be more explicit about why our method should be preferred to goal conditioned RL? we probably also need to compare and forward-reference that comparison

%%SL.12.23: Related work section needs some work, because getting rejected due to comparisons/similarity to prior work is one of the most common things, and the related work section is basically your main defense against that. I suggest:
% discuss skill discovery methods (DIAYN and co) in more detail
% discuss unsupervised meta-RL and relationship to meta-learning more generally
% discuss methods that pretrain and finetune (including finetuning from GCRL)

\cutsectionup
\section{Preliminaries and Problem Setting}
\cutsectiondown

We consider the unsupervised offline reinforcement learning setting, which is defined by a Markov decision process (MDP) along with a dataset $\gD$ of unlabeled transitions. The MDP is fully defined by state space $\gS$, action space $\gA$, a distribution over starting states $p(s_0)$, and a stochastic transition function $p(s_{t+1} \mid s_t, a_t)$. The dataset $\gD$ consists of state-action trajectories of the form $(s_0, a_0, s_1, a_1, \ldots, s_T)$. Note that there is no inherent definition of a reward or goal, and trajectories are not labeled with any form of intent.

In this work, we consider the zero-shot RL problem,
which consists of two phases.
In the first unsupervised pre-training phase,
we aim to train a latent-conditioned policy $\pi(a \mid s, z)$ that captures as diverse behaviors as possible from unlabeled dataset $\gD$, without online environment interactions. 
In the second downstream evaluation phase,
we aim to solve downstream tasks given at test time in a zero-shot manner,
by specifying the latent vectors $z$ that best solve the downstream tasks.
No additional training is allowed once the downstream tasks are revealed.
Each downstream task is defined as a reward function $\eta: \gS \to \sR$,
but we assume access to only a small number of $(s, \eta(s))$ tuples. Tasks share the same environment dynamics. For ease of notation, we denote rewards as functions of state $\eta(s)$, although reward functions may also depend on state-action pairs without loss of generality (i.e., $\eta(s,a)$). 
% We assume that downstream, $N$ samples of a user-given reward function $\eta: \gS \rightarrow \mathbb{R}$ are given in the form of $(s,\eta(s))$ tuples. The objective is to learn a policy that, conditioned on this information, maximizes the expected reward for all downstream tasks. Policies are learned solely from the offline dataset, without online environment interactions. No additional training is done once the downstream task is revealed.

% We assume that a downstream family of evaluation tasks are defined. This family is unknown during training.
% %%SL.1.29: Hmm I don't know if this needs to be a "family" -- we can just say the goal is to pretrain a generalist agent that can infer near-optimal policies for downstream tasks
% During evaluation time, the reward function of the evaluation task is provided to the agent. The objective is to learn a policy which maximizes expected reward for all downstream tasks. Policies are learned solely from the offline dataset, without online environment interactions. No additional training is done once the downstream task is revealed.
% %%SL.1.29: There is a critical aspect of the problem setup that we really need to explain very clearly: the agent gets (s, r) tuples and then does the task. It gets a sample (not "the reward function ... is provided" => samples are provided!). It's important to set this up correctly so that readers fully understand how we evaluate

\begin{figure*}
    \centering
    \includegraphics[width=\textwidth]{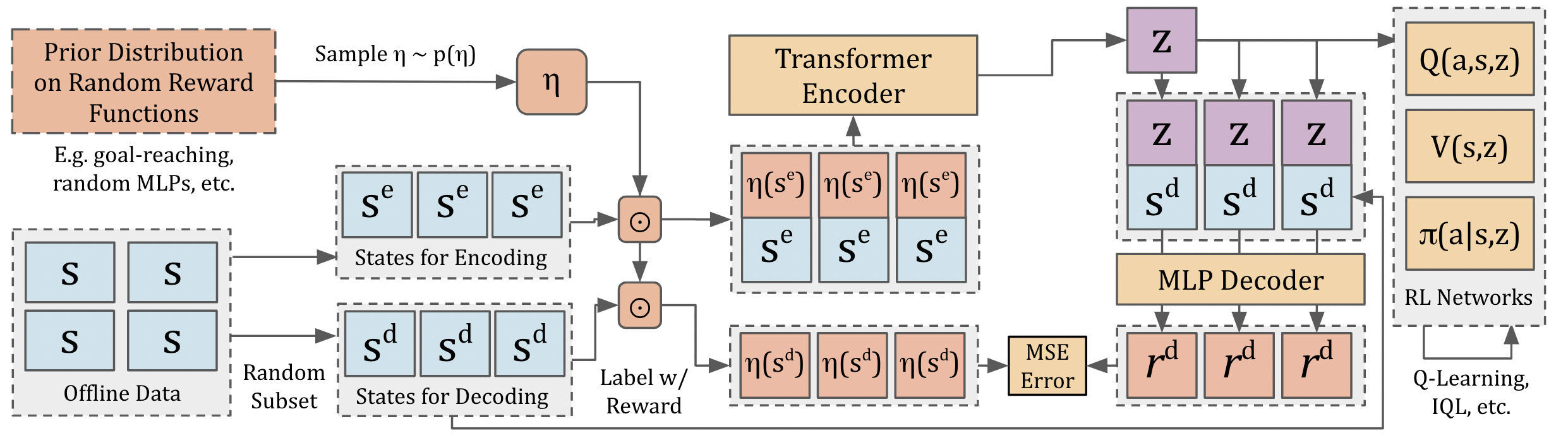}
    %%SL.12.23: This diagram is reasonable, but I might recommend expanding it to more completely illustrate the method -- i.e., actually show the "Downstream use" part so that it's clear that this goes into a policy etc. Otherwise it's kind of illustrating only a piece of the method, and since this is the main technical diagram, it might leave readers a bit confused
    %%SL.1.29: I like the picture, but can we not list the three things it the basis reward box and call it "prior reward distribution" instead?
    %%
    %%
    \vspace{-20pt}
    \caption{\textbf{FRE encodes a reward function by evaluating its output over a random set of data states.} Given a sampled reward function $\eta$, the reward function is first evaluated on a set of random encoder states from the offline dataset. The $(s,\eta(s))$ pairs are then passed into a permutation-invariant transformer encoder, which produces a latent task embedding $z$. A decoder head is then optimized to minimize the mean-squared error between the true reward and the predicted reward on a set of decoder states. The encoder-decoder structure is trained jointly, and $z$ can be utilized for downstream learning of task-conditioned policies and value functions.}
    \label{fig:structure}
    \vspace{-10pt}
\end{figure*}

\cutsectionup
\section{Unsupervised Zero-Shot RL via Functional Reward Encodings}
\cutsectiondown

% If we are out of space, this section is just repeating some things.

Our method, Functional Reward Encoding (FRE), learns to solve arbitrary downstream reward functions by (1) learning to encode diverse random unsupervised rewards into a unified latent space, then (2) training a latent-conditioned policy that can maximize arbitrary rewards from this space.

We begin by developing a neural network encoding over reward functions, connecting to ideas from variational optimization. Our method requires a prior over random reward functions, and we discuss the specific reward prior we use in our experiments, which represents an open design decision for our method. Finally, we propose a practical algorithm that trains such encodings, and uses them to learn zero-shot policies via an off-the-shelf RL algorithm. 

% We begin by developing an encoding that embeds reward functions into a tractable latent space, connecting to ideas from denoising auto-encoders and variational optimization. We then present a set of simple \textit{basis reward families}
% %%SL.1.29: we then discuss the specific reward prior that we use in our experiments, which represents an open design decision for our method
% that roughly span the space of possible reward functions. Finally, we propose a practical algorithm that trains such encodings, and uses them to learn zero-shot policies via an off-the-shelf RL algorithm. 

\cutsubsectionup
\subsection{Functional Reward Encoding}
\cutsubsectiondown

% =============

We present a simple neural network architecture that can encode reward functions according to their \textit{functional form}. The intuition is that a reward function defines a mapping $\eta : \gS \rightarrow \sR$ that can be approximated with samples. Assume that reward functions are distributed according to a prior distribution $p(\eta)$. Under the support of the set of states present in the dataset, any reward function $\eta$ can be represented as a lookup table over the set of state-reward pairs:
\begin{equation}
    L_\eta := \{(s^e, \eta(s^e)) : s^e \in \gD\}
\end{equation}
which defines a corresponding distribution $p(L_\eta)$.

We would like to learn a latent representation $z$ that is maximally informative about $L_\eta$, while remaining maximally compressive. However, as the full state-reward set is intractable due to its size, an approximation must be made.
Our key idea is to make $z$ encoded from a \emph{subset} of state-reward samples to be maximally predictive of \emph{another subset} of state-reward samples, while being as compressive as possible. This can be formulated as the following information bottleneck objective over the structure of $L_\eta^e \to Z \to L_\eta^d$~\citep{tishby2000information, alemi2016deep}:
\begin{align}
I(L_\eta^d; Z) - \beta I(L_\eta^e; Z),
\end{align}
where $L_\eta^e$ and $L_\eta^d$ denote the random variables for the two subsets of $L_\eta$ (with $K$ and $K'$ state-reward tuples, respectively), $Z$ denotes the random variable for the latent vector, and $\beta$ denotes the strength of the compression term.

Since mutual information is generally intractable,
we derive its variational lower bound as follows~\citep{alemi2016deep}:
\begin{align}
    &\quad\; I(L_\eta^d; Z) - \beta I(L_\eta^e; Z) \\
    % &= I(L_\eta^d; Z) - \beta \, \E_{\{s^e\}}[\kldiv{p_\theta(z \mid \{s^e\})}{p_\theta(z)}] \\
    % &\geq I(L_\eta^d; Z) - \beta \, \E_{\{s^e\}}[\kldiv{p_\theta(z \mid \{s^e\})}{u(z)}] \\
    % \begin{split} \label{eq:objective}
    % &\geq \E_{\{s^e\}, \{s^d\}, \eta, z \sim p_\theta(z \mid \cdot)}\Bigg[\sum_{k=0}^{K-1} \log q_\theta(\eta(s^d_k) \mid s^d_k, z) \\
    % &\qquad -\beta \kldiv{p_\theta(z \mid \{s^e\})}{u(z)}\Bigg] + (\mathrm{const}),
    &= I(L_\eta^d; Z) - \beta \, \E[\kldiv{p_\theta(z \mid L_\eta^e)}{p_\theta(z)}] \\
    &\geq I(L_\eta^d; Z) - \beta \, \E[\kldiv{p_\theta(z \mid L_\eta^e)}{u(z)}] \\
    \begin{split} \label{eq:objective}
    &\geq \E_{\eta, L_\eta^e, L_\eta^d, z \sim p_\theta(z \mid L_\eta^e)}\Bigg[\sum_{k=1}^{K'} \log q_\theta(\eta(s^d_k) \mid s^d_k, z) \\
    &\qquad -\beta \kldiv{p_\theta(z \mid L_\eta^e)}{u(z)}\Bigg] + (\mathrm{const}),
    \end{split}
\end{align}
where
we slightly abuse the notation by using $L_\eta^e$ to denote both the random variable and its realization,
such as
\begin{equation} 
p_\theta(z \mid L_\eta^e) = p_\theta(z \mid s^e_1,\eta(s^e_1),s^e_2,\eta(s^e_2), \ldots, s^e_K,\eta(s^e_K)), \nonumber
\end{equation}
and $u(z)$ is an uninformative prior over $z$, which we define as the unit Gaussian.
% Here we make use of the fact that $L^d$ is a random variable of state-reward tuples $(s, \eta(s))$, and the marginal $p(s)$ under $\gD$ is equivalent for any reward function. Thus, $p((s, \eta(s)))$ can be factorized into $p(\eta(s)|s)p(s)$ which results in the expectation over conditionals of $\mathbb{E}_{\{s^d\}}[ q_\theta(\eta(s^d) \mid s^d,z) ]$ in Equation \ref{eq:objective}.
Here, we make use of the fact that $\log q_\theta(L_\eta^d \mid z) = \sum_k \log q_\theta(s_k^d, \eta(s_k^d) \mid z) = \sum_k \log q_\theta(\eta(s_k^d) \mid s_k^d, z) + (\mathrm{const})$.

Training an FRE requires two neural networks, 
\begin{align}
    & \text{Encoder:} \; p_\theta(z \mid s^e_1,\eta(s^e_1),s^e_2,\eta(s^e_2), \ldots, s^e_K,\eta(s^e_K)), \\
    & \text{Decoder:} \; q_\theta(\eta(s^d) \mid s^d,z).
\end{align}
which are jointly optimized towards the objective described in Equation \ref{eq:objective}. FRE therefore learns a minimal latent representation $z$ that is maximally informative about $L_\eta$, which may be used in downstream offline RL algorithms.

The FRE method is similar to a denoising auto-encoder \cite{vincent2008extracting} trained on $(s,\eta(s))$ pairs sampled from an arbitrary reward function $\eta$ over $s^e, s^d \sim \gD$, as well as that of neural processes \cite{garnelo2018neural, garnelo2018conditional, kim2019attentive} in that we aim to map a context set of reward-state pairs to a functional output. The main difference is that both denoising auto-encoders and neural processes utilize a determinstic encoder, and we instead use a probabilistic encoder with an information penalty. Additionally, FRE uses a fixed number of samples $K$ while neural processes generally use a variable number.

\textbf{Practical Implementation.} In our experiments, the encoder $p_\theta(z \mid \cdot)$ is implemented as a permutation-invariant transformer \cite{vaswani2017attention}. $K$ encoder states are sampled uniformly from the offline dataset, then labeled with a scalar reward according to the given reward function $\eta$. The resulting reward is discretized according to magnitude into a learned embedding token space. The reward embeddings and states are then concatenated as input to the transformer. Positional encodings and causal masking are not used, thus the inputs are treated as an unordered set. The average of the final layer representations is used as input to two linear projections which parametrize the mean and standard deviation of Gaussian distribution $p_\theta(z \mid \cdot)$. 

The decoder $q_\theta(\eta(s) \mid s,z)$ is implemented as a feedforward neural network. Crucially, the states sampled for decoding are different than those used for encoding. The encoding network makes use of the entire set of $(s_{1..K},\eta(s_{1..K}))$ pairs, whereas the decoder independently predicts the reward for each state, given the shared latent encoding $z$. We train both the encoder and decoder networks jointly, minimizing mean-squared error between the predicted and true rewards under the decoding states.

\cutsubsectionup
\subsection{Random Functions as a Prior Reward Distribution}
\cutsubsectiondown

%%SL.1.29: maybe we can call it prior reward distribution?

An FRE encoding depends on (1) an offline dataset of trajectories, and (2) a distribution of reward functions. While trajectories are provided, we do not know ahead of time the downstream reward functions. Thus, we aim to craft a relatively uninformative but diverse prior over reward functions. 

The specific choice of prior reward distribution is a design choice. While completely random functions lead to incompressible representations (as per the No Free Lunch theorem \cite{wolpert1997no}), more structured distributions can lead to robust representations and generalization. Reasonable choices should broadly span the space of possible downstream tasks, while remaining domain-agnostic.

In our implementation, we found that a reasonable yet powerful prior distribution can be constructed from a mixture of \textit{random unsupervised functions}. The particular mixture we use consists of random singleton functions (corresponding to ``goal reaching'' rewards), random neural networks (MLPs with two linear layers), and random linear functions (corresponding to ``MLPs'' with one linear layer). This provides both a degree of structure and a mixture of high frequency (singletons) and low frequency (linear) functions, with the MLPs serving as an intermediate function complexity. A uniform mixture of the three function classes are used during training. We study these choices further in Section~\ref{sec:rewards}.

If we have privileged knowledge about the downstream tasks, we can adjust the prior reward distribution accordingly, as we will discuss in Section~\ref{sec:improvement}.

\cutsubsectionup
\subsection{Offline RL with FRE}
\cutsubsectiondown

To close the loop on the method, we must learn an FRE-conditioned policy that maximizes expected return for tasks within the prior reward distribution. Any off-the-shelf RL algorithm can be used for this purpose. The general pipeline is to first sample a reward function $\eta$, encode it into $z$ via the FRE encoder, and optimize $\pi(a \mid s,z)$. 

At each training iteration, a batch of state-action pairs $(s,a)$ are selected from the offline dataset. Additionally, a batch of reward functions $\eta$ are also sampled from the prior reward distribution. Each reward function is evaluated on $K$ encoding states from the offline dataset. The resulting $(s^e, \eta(s^e))$ context pairs are then passed into the FRE encoder to produce a latent representation $z$.

The latent representation $z$ can then be used for RL training. The RL components (Q-function, value function, and policy) are all conditioned on $z$. The sampled reward function $\eta$ is used to calculate rewards during training. A standard Bellman policy improvement step using FRE looks like:
\begin{equation}
    Q(s,a,z) \gets \eta(s) + \E_{s' \sim p(s'|s, a)}\left[\max_{a' \in \gA} Q(s',a',z)\right],
\end{equation}

\textbf{Practical Implementation.} In our experiments, we use implicit Q-learning \cite{kostrikov2021offline} as the offline RL method to train our FRE-conditioned policy. This is a widely used offline RL algorithm that avoids querying out-of-distribution actions. 

% The resulting algorithm learns to optimize a basis set of reward families over a set of offline trajectories. During test time, novel downstream reward families can be provided, and policies are executed in a zero-shot manner.

% In the IQL formulation, separate Q- and value functions are trained. The value function is updated with an expectile loss that can be seen as a form of soft maximization. Policy extraction is performed via advantage-weighted regression \cite{peng2019advantage}. Advantages are computed as $A(s,a,z) = Q(s,a,z) - V(s,z)$. 

% We find that a strided training scheme leads to the most stable performance. In the strided scheme, a warmup period first takes place in which the FRE encoder is trained with gradients from the decoder. During this time, the RL components are not trained. After the warmup period has ended, the RL networks are then trained using the frozen encoder. Gradients are not passed through to the encoder network from the RL components. The stationary of the mapping from $\eta$ to $z$ during training is important to correctly estimate multi-task Q values using TD learning. We summarize our training procedure of FRE in \Cref{alg:fre}.
We find that a strided training scheme leads to the most stable performance. In the strided scheme, we first only train the FRE encoder with gradients from the decoder (\Cref{eq:objective}). During this time, the RL components are not trained. After the encoder loss converges, we freeze the encoder and then start the training of the RL networks using the frozen encoder's outputs. In this way, we can make the mapping from $\eta$ to $z$ stationary during policy learning, which we found to be important to correctly estimate multi-task Q values using TD learning. We summarize our training procedure of FRE in \Cref{alg:fre}.

\begin{algorithm}[t!]
   \caption{Functional Reward Encodings (FRE)}
   \label{alg:fre}
%    \STATE {\bfseries Input:} Offline dataset $\gD$ of trajectories $(s_{0..T},a_{0..T-1})$. 
%    \STATE Random reward function distribution $p(\eta)$.  
%    \STATE \textbf{Begin:}
%    \FOR{$i=0$ {\bfseries to} $I$}

%         \STATE \textit{\# Select state-actions and random reward func.}
%         \STATE Sample $N$ transitions $(s,a,s') \sim \gD$.
%         \STATE Sample $N$ reward functions $\eta \sim p(\eta)$.

%         \STATE \textit{\# Encode each reward func into FRE representation.}
%         \STATE Sample $K$ pairs $(s, \eta(s)) \sim \gD$
%         \STATE Encode into latent vector $z \sim p_\theta(s_{0..k},r_{0..k})$.

%         \STATE \textit{\# If warmup period, train encoder and decoder.}
%         % \STATE $q_\theta(r|s,w) \leftarrow r_w(s)$
%         \STATE $q_\theta(r \mid s,z) \leftarrow \eta(s)$

%         %%SL.12.23: I wonder if it would be better if the pseudocode presents the generic version of the algorithm that is not specific to any particular RL method
%         \STATE \textit{\# Else, train RL components.}
%         \STATE $V(s,z) \leftarrow V(s,z) \propto \mathrm{exptl}(Q(s,a,z) - V(s,z))$
%         \STATE $Q(s,a,z) \leftarrow \eta(s) + \gamma V(s',z)$
%         \STATE $\pi(a \mid s,z) \leftarrow \exp(Q(s,a,z) - V(s,z))$
%         %%SL.12.23: the \leftarrow symbol is a very imprecise way to describe what happens, perhaps we should instead introduce symbols for the losses and say we update with gradients on those losses?
        
%    \ENDFOR
% \end{algorithmic}
\small
\begin{algorithmic}
   \STATE \textbf{Input:} unlabeled offline dataset $\gD$, distribution over random unsupervised reward functions $p(\eta)$.  
   \STATE \textbf{Begin:}
   \STATE \emph{\# Train encoder}
   \WHILE{not converged}
        \STATE Sample reward function $\eta \sim p(\eta)$
        \STATE Sample $K$ states for encoder $\{s^e_k\} \sim \gD$ %and label $\{(s^e_k, \eta(s^e_k))\}$
        \STATE Sample $K'$ states for decoder $\{s^d_k\} \sim \gD$
        % \STATE Sample $N$ reward functions $\{\eta_n\} \sim p(\eta)$.
        % \STATE Sample $NK$ states for encoder $\{s^e_{n, k}\} \sim \gD$ %and label $\{(s^e_k, \eta(s^e_k))\}$
        % \STATE Sample $NK'$ states for decoder $\{s^d_{n, k}\} \sim \gD$
        \STATE Train FRE by maximizing \Cref{eq:objective}
   \ENDWHILE
   \STATE \emph{\# Train policy}
   \WHILE{not converged}
        \STATE Sample reward function $\eta \sim p(\eta)$
        \STATE Sample $K$ states for encoder $\{s^e_k\} \sim \gD$
        \STATE Encode into latent vector $z \sim p_\theta(\{(s^e_k, \eta(s^e_k))\})$
        % \STATE Sample $N$ reward functions $\{\eta_n\} \sim p(\eta)$.
        % \STATE Sample $NK$ states for encoder $\{s^e_{n, k}\} \sim \gD$ %and label $\{(s^e_k, \eta(s^e_k))\}$
        % \STATE Encode into latent vectors $z_n \sim p_\theta(\{s^e_{n, k}, \eta(s^e_{n, k})\})$.
        \STATE Train $\pi(a|s, z)$, $Q(s, a, z)$, $V(s, z)$ using IQL with $r = \eta(s)$
        
   \ENDWHILE
\end{algorithmic}
\end{algorithm}

\cutsectionup
\section{Experiments} \label{sec:experiments}
\cutsectiondown
%%SL.12.23: I recommend staritng the experiments section with a discussion of the main goals of the experiments and the questions the experiments section is aiming to answer. What you have below looks like experiment setup, which probably should be placed in a separate \paragraph or subsection labeled as such

In the following section, we present a series of experiments confirming the effectiveness of FRE as an unsupervised zero-shot RL method. Results are presented on standard offline RL benchmarks: the ExORL benchmark for learning from unsupervised data \cite{yarats2022don} and variants of the AntMaze and Kitchen environments from D4RL \cite{fu2020d4rl} adapted for evaluating multi-task and goal-conditioned policies. We evaluate on tasks chosen to be representative of the main challenges of each domain, extending the standard tasks whenever possible.
%%SL.1.29: tweaked the above, double-check that it's right plz

\begin{figure*}
    \centering
    \includegraphics[width=0.9\textwidth]{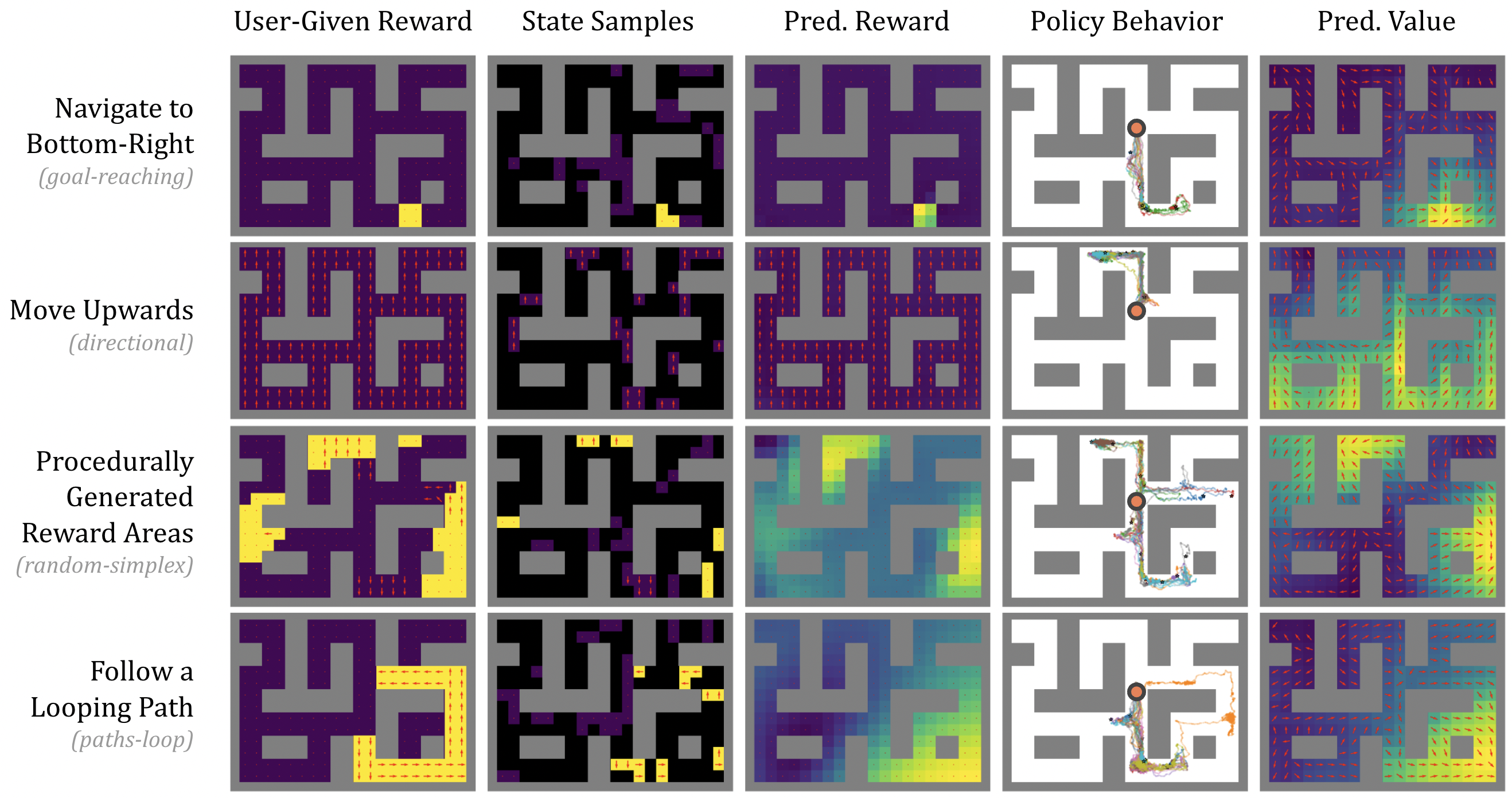}
    %%SL.1.29: Can we do more to improve readability of this figure? I think the idea makes a ton of sense, but bits of it are hard to see: (i) The purple (?) dots representing the samples are very hard to see; (ii) the trajectory lines are pretty thin, hard to make out in the 4th column; I also can't tell at all from this if this policy is good or bad -- any way to better indicate if it is going to the right place?; (iii) it's unclear what the yellow blotches vs purple dots for "state samples" represent. Additionally, maybe it would be good to take out the "directional movements" row? It's very hard to tell what is going on there, and it's also not clear why that's really relevant for the maze

    \caption{\textbf{After unsupervised pretraining, FRE can solve user-specified downstream tasks without additional fine-tuning.} Shown above are examples of reward functions sampled from various evaluations in AntMaze. \textbf{Columns:} 1) True reward function projected onto maze. 2) Random states used for encoding shown in non-black. 3) Reward predicted by decoder network. 4) Behavior of FRE policy conditioned on latent encoding. Agents start at the red dot. 5) Visualization of predicted value function. }
    \label{fig:antmaze}
\end{figure*}

\begin{table*}
  \centering
  \begin{tabular}{l|llllll}
    \toprule
    % \cmidrule(r){1-2}
    Eval Task   & FRE & FB & SF & GC-IQL & GC-BC & OPAL-10$^1$ \\
    \midrule
ant-goal-reaching & \boldmath{$48.8 \pm 6$} & $0.0 \pm 0$ & $0.4 \pm 2$ & \boldmath{$40.0 \pm 14$} & $12.0 \pm 18$ & $19.4 \pm 12$ \\
ant-directional & \boldmath{$55.2 \pm 8$} & $4.8 \pm 14$ & $6.5 \pm 16$ & - & - & $39.4 \pm 13$ \\
ant-random-simplex & \boldmath{$21.3 \pm 4$} & $9.7 \pm 2$ & $8.5 \pm 10$ & - & - & \boldmath{$27.3 \pm 8$} \\
% ant-path-loop & \boldmath{$67.2 \pm 36$} & $46.6 \pm 40$ & $13.6 \pm 16$ & $35.9 \pm 5$ & $9.6 \pm 1$ & $44.4 \pm 22$ \\
% ant-path-edges & $60.0 \pm 17$ & $23.5 \pm 25$ & $2.2 \pm 5$ & $31.7 \pm 3$ & $23.1 \pm 10$ & \boldmath{$85.0 \pm 10$} \\
% ant-path-center & \boldmath{$64.4 \pm 38$} & \boldmath{$70.3 \pm 37$} & $39.4 \pm 27$ & $35.2 \pm 5$ & $21.2 \pm 9$ & $58.1 \pm 36$ \\
ant-path-loop & \boldmath{$67.2 \pm 36$} & $46.6 \pm 40$ & $13.6 \pm 16$ & - & - & $44.4 \pm 22$ \\
ant-path-edges & $60.0 \pm 17$ & $23.5 \pm 25$ & $2.2 \pm 5$ & - & - & \boldmath{$85.0 \pm 10$} \\
ant-path-center & \boldmath{$64.4 \pm 38$} & \boldmath{$70.3 \pm 37$} & $39.4 \pm 27$ & - & - & $58.1 \pm 36$ \\
\midrule
antmaze-all & \boldmath{$52.8 \pm 18.2$} & $25.8 \pm 19.8$ & $11.8 \pm 12.6$ & - & - & $45.6 \pm 17.0$ \\

\midrule

exorl-walker-goals & \boldmath{$94 \pm 2$} & $58 \pm 30$ & \boldmath{$100 \pm 0$} & \boldmath{$92 \pm 4$} & $52 \pm 18$ & \boldmath{$88 \pm 8$} \\
exorl-cheetah-goals & $58 \pm 8$ & $1 \pm 2$ & $0 \pm 0$ & \boldmath{$100 \pm 0$} & $14 \pm 6$ & $0 \pm 0$ \\
exorl-walker-velocity & $34 \pm 13$ & \boldmath{$64 \pm 1$} & $38 \pm 4$ & - & - & $8 \pm 0$ \\
exorl-cheetah-velocity & $20 \pm 2$ & \boldmath{$51 \pm 3$} & $25 \pm 3$ & - & - & $17 \pm 8$ \\
\midrule
exorl-all & \boldmath{$51.5 \pm 6.3$} & $43.4 \pm 9.1$ & $40.9 \pm 1.9$ & - & - & $28.2 \pm 4.0$ \\
    \midrule

kitchen & \boldmath{$66 \pm 3$} & $3 \pm 6$ & $1 \pm 1$ & \boldmath{$59 \pm 4$} & $35 \pm 9$ & $26 \pm 16$ \\
\midrule
\midrule

all & \boldmath{$57 \pm 9$} & $24 \pm 12$ & $18 \pm 5$ & - & - & $33 \pm 12$ \\

\bottomrule
    
  \end{tabular}
  \vspace{0.5cm}
\caption{\textbf{Offline zero-shot RL comparisons on AntMaze, ExORL, and Kitchen.} FRE-conditioned policies match or outperform state-of-the-art prior methods on many standard evaluation objectives including goal-reaching, directional movement, and structured locomotion paths. FRE utilizes only 32 examples of (state, reward) pairs during evaluation, while the FB and SF methods require 5120 examples to be consistent with prior work. Results are normalized between 0 and 100. \\ \\
$^1$OPAL is a skill discovery method and does not have zero-shot capabilities. Thus, we compare to a privileged version where the agent evaluates 10 skills in the downstream task with \emph{online} rollouts, and selects the one with the highest performance.
}

% }
\label{table:benchmark}
\end{table*}

\begin{figure}
  \centering
    \includegraphics[width=0.5\textwidth]{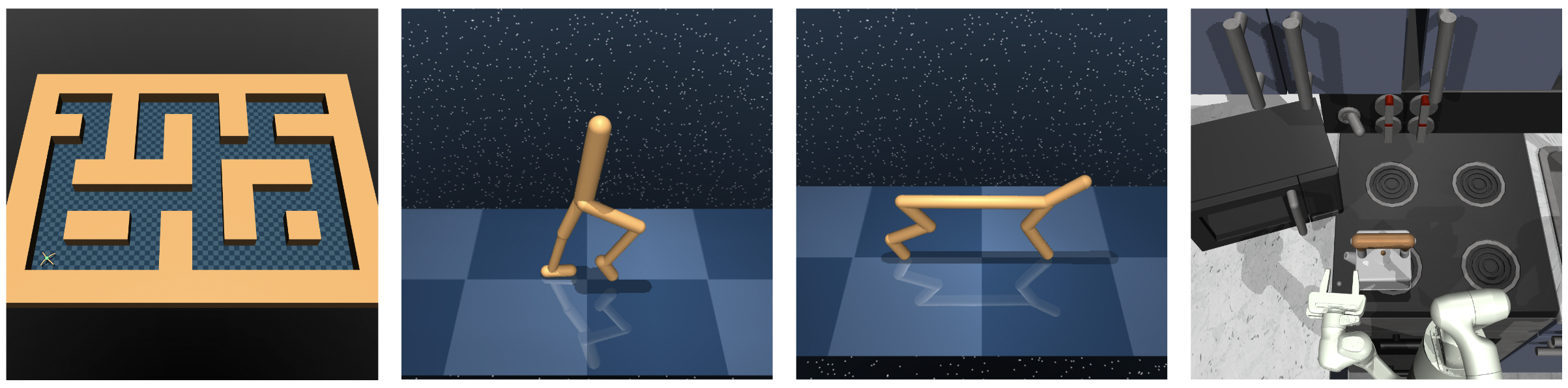}
\vspace{-15pt}
\caption{Evaluation domains: AntMaze, ExORL, and Kitchen.}
\label{fig:tasks}
\vspace{-15pt}
\end{figure}

The \textbf{ExORL} dataset is a standard collection of offline data for RL, consisting of trajectories sampled by an exploratory policy on DeepMind Control Suite \cite{tassa2018deepmind} tasks. We consider the \textit{walker} and \textit{cheetah} domains, in accordance with \cite{touati2022does}. To examine zero-shot capabilities, we examine transfer to the standard reward functions consisting of forward/backward velocity, along with goal-reaching to random states in the dataset.

\textbf{AntMaze} is a benchmark task where an 8-DoF Ant robot must be maneuvered around a maze.
%%SL.1.29: maybe more relevant to say the number of state dimensions rather than DoFs?
We use the most challenging offline AntMaze dataset from D4RL \cite{fu2020d4rl} under the \textit{antmaze-large-diverse-v2} name. Data trajectories consist of walking paths within the maze. We consider four natural families of tasks: (1) a \textit{goal-reaching} task where the robot must navigate to various locations in the maze, (2) a \textit{directional} task which involves moving in a given $(x,y)$ direction, (3) a \textit{random-simplex} task which assigns reward based on a procedural noise generator, and (4) a set of three hand-designed \textit{path} tasks involving navigating to the edges of the map, moving in a loop, and following a central corridor.

% %%SL.1.29: the data consists of walking paths between random start and end points in the maze
% We chose AntMaze as a setting where the true reward function is less obvious, and may consist of various locomotion goals.
% %%SL.1.29: Within this domain, we can construct a variety of downstream locomotion tasks to evaluate our method's ability to recover diverse skills? then top down sturcture like: we consider four types of tasks...
% We consider a downstream \textit{goal-reaching} task where the robot must navigate to various locations in the maze. The \textit{directional} task involves moving in a given $(x,y)$ direction, with dense reward given based on movement at the desired velocity. The \textit{random-simplex} task defined reward functions based on a procedural noise generator \cite{olano2002real}. To further examine zero-shot generalization, we additionally define a set of three hand-designed test tasks in the AntMaze domain. The test tasks involve navigating to the edges of the map, moving in a loop, and following a directional path.

\textbf{Kitchen} is a D4RL environment where a robot hand must accomplish various objectives in an interactive environment (opening a microwave, flipping a light switch, etc.). To extend Kitchen into a multi-task setting, we evaluate on reward functions corresponding to each of the 7 standard subtasks in the environment.

%%SL.1.29: explain that these require moving objects, i.e., its not a goal state but a goal object pose

%%SL.1.29: the word "expert" is pretty loaded here, b/c they are not necessarily expert for a particular task, they just do random tasks -- can we make this clearer?

%%SL.1.29: It would be really good to reference a figure that shows visualizations of these envs. While domain experts might not need this, anyone who is not familiar with the environments will really appreciate being able to see some visualziations

\textbf{Prior Reward Distribution.} We utilize the same prior reward distribution for training FRE agents on each domain. Specifically, we consider a mixture of three random unsupervised function types, each with progressively higher complexity. The first are singleton \textbf{goal-reaching} rewards corresponding to a reward of -1 for every timestep where the goal has not been reached, and 0 otherwise. Goals are sampled randomly from the dataset. The second are \textbf{random linear functions}, defined as the inner product between a uniformly random vector and the current state. We find that biasing towards simple functions is a useful prior, which can be achieved via a sparse mask over the vector. The final family is \textbf{random MLPs}, which are implemented as random initializations of 2-layer MLPs. MLPs with sufficient size are universal function approximators \cite{hornik1989multilayer}, thus this family broadly covers possible downstream rewards.

\begin{figure*}
  \centering
    \includegraphics[width=1\textwidth]{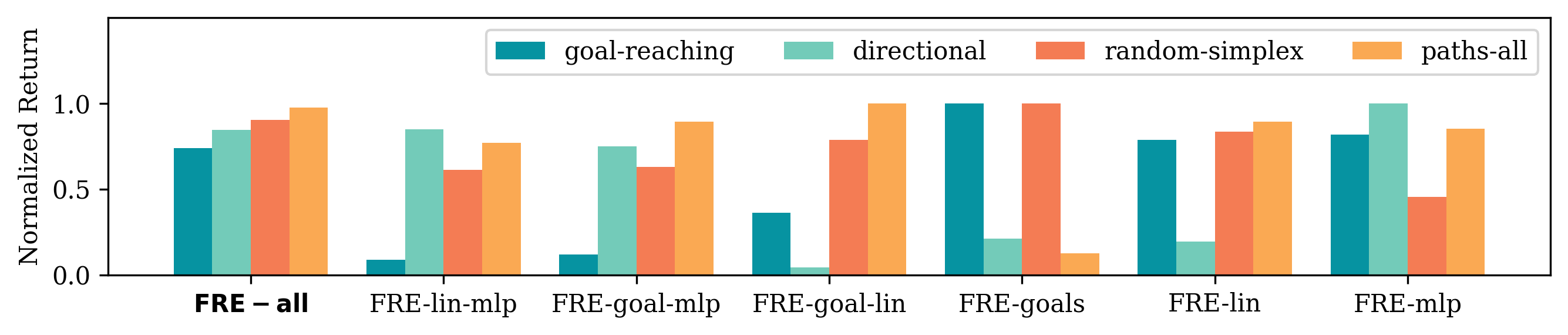}
\vspace{-20pt}
\caption{\textbf{The general capabilities of a FRE agent scales with diversity of random functions used in training.} FRE-all represents an agent trained on a uniform mixture of three random reward families, while each other column represents a specific agent trained on only a subset of the three. The robust FRE-all agent displays the largest total score, and competitive performance among all evaluation tasks, showing that the FRE encoding can combine reward function distributions without losing performance. }
\label{fig:transfer}
\vspace{-10pt}
\end{figure*}

\cutsubsectionup
\subsection{Do FRE encodings trained on random reward functions zero-shot transfer to unseen test tasks?}
\cutsubsectiondown

Figure~\ref{fig:antmaze} illustrates how FRE generalizes from samples of randomly-annotated AntMaze states, both in terms of the decoded reward and the resulting policy and estimated value function. In all cases, the value function correctly captures the approximate expected returns from each position. The executed policy generally maximizes the value function, although some trajectories fail when encountering out-of-distribution states, which is a common challenge in offline RL~\cite{kumar2020conservative}. Thus, we can conclude that FRE encodings present a simple yet reasonable way of solving downstream tasks without additional training.

% We then examine the outputs of the various FRE networks. The decoder network output is presented as ``Pred. Reward''. While this output is not actually used at test time, we can examine these predictions to understand how the encoder is capturing information about the reward function. Next, the behavior of an executed FRE policy is plotted as a series of twenty trajectories. The value function of the policy is also shown. 

% Thus, we can conclude that FRE encodings trained on diverse randomized reward functions can solve downstream tasks without additional training.
%%SL.1.29: if are running low on space, it would be reasonable to heavily curtail the description of the figure above, since much of it will be redundant with the figure caption

\cutsubsectionup
\subsection{How does FRE perform on zero-shot offline RL benchmarks, compared to prior methods?}
\cutsubsectiondown

We now examine the performance of our FRE agent on new downstream tasks, and compare with state-of-the-art prior methods in unsupervised RL. The comparisons include:

% We examine the performance of two varieties of RARE-IQL agents, along with state-of-the-art prior methods when applied to multi-task offline learning benchmarks. The comparisons comprise:

%%SL.1.29: for prior methods, can you put citations on them? if it's unclear which specific prior method it is, ok to put multiple citations
\begin{itemize}[itemsep=0pt, topsep=0pt, leftmargin=10pt]
    \item \textbf{FRE}, our method.
    %%SL.1.29: suggestion on phrasing: instead of calling it "three basis reward families" call it "our proposed reward distribution" (basically, compartmentalize, leave the "three" to that paragraph the describes the distribution)
    \item \textbf{Forward-Backward (FB)} method \cite{touati2021learning}, a state-of-the-art zero-shot RL method that jointly learns a pair of representations that represent a family of tasks and their optimal policies.
    \item \textbf{Successor Features (SF)} \citep{barreto2017successor, borsa2018universal}, which utilize a set of pre-trained features to approximate a universal family of reward functions and their corresponding policies.
    \item \textbf{Goal-Conditioned IQL (GC-IQL)} \cite{kostrikov2021offline}, a representative goal-conditioned RL method. GC-IQL is a variant of IQL that uses hindsight relabeling to learn goal-reaching policies.
    \item \textbf{Goal-Conditioned Behavioral Cloning (GC-BC)}, a simple offline RL method that learns goal-reaching policies by mimicking trajectories that reach goals in the dataset.
    \item \textbf{OPAL} \cite{ajay2020opal}, a representative offline unsupervised skill discovery method where latent skills are learned by auto-encoding trajectories. 
\end{itemize}

\begin{table}

  \newcommand{\cmark}{\ding{51}}%   % 51: thinner, 52: thicker
  \newcommand{\xmark}{\ding{55}}%
  \newcommand{\Y}{{\color[HTML]{38761D}\cmark}\xspace}%
  \newcommand{\N}{{\color[HTML]{990000}\xmark}\xspace}%

  \scalebox{0.9}{
  \centering
  \begin{tabular}{l|lllll}
    \toprule
    % \cmidrule(r){1-2}
       & FRE & FB & SF & GCRL & OPAL\\
    \midrule
    Zero-Shot & \Y & \Y & \Y & \Y & \N \\
    Any Reward Func. & \Y & \Y & \N & \N & \N\\
    No Linear Constraint & \Y & \N & \N & \Y & \Y\\
    Learns Optimal Policies & \Y & \Y & \Y & \Y & \N \\

    \bottomrule
  \end{tabular}
  }

  \caption{\textbf{FRE unifies prior methods in capabilities}. OPAL does not have zero-shot capabilities and learns via BC rather than Q-learning. GCRL and SF both limit reward function families to goal-reaching or linear functions, respectively. FB can learn to solve any reward function, but requires a linearized value function.}

\label{table:comparison}
\end{table}

% In all experiments, each method receives a sample of states with reward labels to infer the task. Prior work refers to this setting as ``zero shot''~\citep{touati2022does}, in the sense that although each method receives state-reward tuples for inference, no additional updates to the models are performed for the downstream task. This also resembles past work on meta-reinforcement learning, though without access to meta-training tasks~\citep{duan2016rl}.

% In all experiments, the performance of the learned policies is evaluated in a zero-shot manner.
%%SL.1.29: let's be careful with "zero shot" -- we do give it data in the form of state-reward tuples that are used for inference. Can we say maybe "In all experiments, each method receives a sample of states with reward labels to infer the task. Prior work refers to this setting as ``zero shot''~\citep{}, in the sense that although each method receives state-reward tuples for inference, no additional updates to the models are performed for the downstream task. This also resembles past work on meta-reinforcement learning, though without access to meta-training tasks~\citep{}."
All methods are evaluated using a mean over twenty evaluation episodes,
%%SL.1.29: maybe mention here or elsewhere what the +/- represents (variance? standard error? something else?)
and each agent is trained using five random seeds, with the standard deviation across seeds shown. FRE, GC-IQL, and GC-BC are implemented within the same codebase and with the same network structure. FB and SF are based on DDPG-based policies, and are run via the code provided from \cite{touati2022does}. For the SF comparisons, we follow prior work \cite{touati2022does}
%%SL.1.29: can you cite that prior work too? presumably it's not just pathak, since that doesn't have SF
and learn features using ICM~\cite{pathak2017curiosity}, which is reported to be the strongest method in the ExORL Walker and Cheetah tasks~\citep{touati2022does}. OPAL is re-implemented in our codebase.

Table \ref{table:benchmark} shows that FRE matches or outperforms state-of-the-art baselines on AntMaze and the ExORL benchmark. Especially on goal-reaching tasks, the FRE agent is able to considerably outperform SF-based baselines, and matches goal-specific RL methods. The same FRE agent is able to solve a variety of other tasks, such as directional movement, random simplex rewards, and unique test tasks.
%%SL.1.29: it would be good if the names of tasks above match the names you used when introducing the env previously
Thus FRE is shown to be a competitive method for simple zero-shot unsupervised RL.

Similarly to FRE, FB and SF train a universal agent that maximizes unsupervised reward functions. They rely on linearized value functions to achieve generalization, whereas FRE learns a shared latent space through modeling a reward distribution. Note that FB/SF rely on linear regression to perform test time adaptation, whereas FRE uses a learned encoder network. To be consistent with prior methodology, we give these methods 5120 reward samples during evaluation time (in comparison to only 32 for FRE). Even with considerably fewer samples, FRE is competitive with a strong FB method across the board, and greatly outperforms on \textit{goal-reaching}, \textit{ant-directional}, and \textit{kitchen}. OPAL is considered as an offline unsupervised skill discovery method that also learns from offline data. Since OPAL does not solve the problem of understanding a reward function zero-shot, we compare to a version with privileged execution based on online rollouts.
%%SP: I think it'd be great to add one sentence about the comparison between FB and OPAL here (like the below), but not sure what we should write...
Despite OPAL's use of privileged online rollouts, however, the results suggest that FRE outperforms OPAL in general, often by a significant margin.

\cutsubsectionup
\subsection{What are the scaling properties of FRE as the space of random rewards increases?} \label{sec:rewards}
\cutsubsectiondown

One desirable property of FRE is that disparate reward families can be jointly encoded. We examine if encoding combinations of random reward families lead to stronger performance (due to generalization) or weaker performance (due to limited network capacity, forgetting, etc). We train FRE agents on all possible subsets of the random reward forms. All agents are given the same training budget, e.g. FRE-all has a third of the goal-reaching tasks of FRE-goals.

Table \ref{table:transfer} highlights that the FRE-all agent trained on all reward types displays the highest total score, and competitive performance among all evaluation tasks.
This result suggests that the performance of FRE scales smoothly as we use more diverse types of rewards,
thanks to our scalable architecture for the permutation-invariant encoder.

% It remains important to define training tasks that reasonably cover the space of evaluation tasks. The random-mlp set is demonstratably more effective in generalization to the unseen test tasks provided. Thus, if knowledge about the set of desired downstream behaviors is known, we recommend to define a range of reward families that cover the space of desired behaviors. As shown above, multiple reward families can be easily combined, thus it is better to have too large a training space than too small.

\cutsubsectionup
\subsection{Can prior domain knowledge be used to increase the specificity of the FRE encoding?} \label{sec:improvement}
\cutsubsectiondown

\begin{figure}
    \centering
    \includegraphics[width=0.45\textwidth]{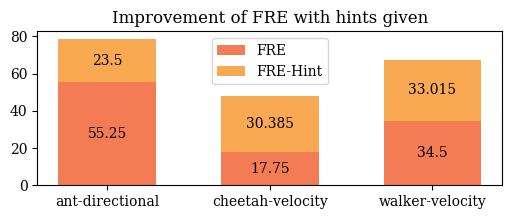}
    \vspace{-10pt}
    \caption{By augmenting the random reward families with specific reward distributions, FRE can utilize domain knowledge without algorithmic changes.}
    \label{fig:hints}
    \vspace{-15pt}
\end{figure}

Oftentimes, some information is known about the distribution of downstream tasks beforehand. FRE can straightforwardly utilize this info by augmenting the prior reward functions with a more specific form, e.g. random functions that depend only on XY positions or velocity.

% with a more specific distribution. As an example, if we know that downstream rewards are all a function of a subset of the state (e.g., XY positions) then we can define random reward functions over that subspace. If we know that downstream tasks are of a certain functional form (e.g., locomotion in a direction), then we can plug in that distribution of reward functions as a basis family during training.
%%SL.1.29: perhaps it would be good somewhere to relate this to the meta-RL problem statement, which can be seen as one extreme of this spectrum? (though not sure where to do that or how essential that is, but some readers might appreciate it and it might avoid some annoying reviewer questions like "how is this different from meta-RL?")

% Figure \ref{fig:hints} highlights the effect of utilizing prior knowledge to train FRE policies.  The benefits of prior knowledge are twofold: first, the encoding network can better compress reward functions in the subspace, and second, the policy can encounter relevant reward functions at a higher rate.

Figure \ref{fig:hints} highlights the universality of FRE by utilizing it as-is as a \textit{multi-task RL} method, where policies are optimized over \textit{known} task distributions. No changes to the neural network architecture or algorithm are necessary. Thus FRE has an advantage in generality over multi-task methods such as GCRL which are limited to a specific form of task. FRE allows us to bypass the need for domain-specific task labeling, as approximate latent encodings of each task are learned in a unsupervised manner.

\cutsectionup
\section{Discussion}
\cutsectiondown

%%SL.12.23: It's always good to end the paper on some inspiring discussion, eg, conclude by talkign about the future work avenues that are opened by this work (which can follow naturally from discussion of limitations)
This work describes a general unsupervised method for use in zero-shot offline reinforcement learning. We first introduce a \textit{functional} encoding for reward functions, allowing us to learn a universal latent representation of tasks. When trained over only random reward functions, FRE-conditioned policies are able to generalize to novel downstream rewards. FRE displays competetive performance on goal-reaching and multi-task unsupervised settings, using a single agent.

\textbf{Limitations.} While FRE provides a flexible and nonlinear policy inference for new tasks, it relies on a hand-specified prior reward distribution. Our specific reward distribution is relatively simple, consists of a mixture of various random functions. However, this choice is somewhat ad hoc, and while we empirically observe that it works well for many benchmark tasks, it remains unclear if there is an optimal and general choice of reward priors. The formulation presented in this work requires offline data, and extending FRE to the online setting is a promising direction for future work.

% The random reward functions presented in this work are simple choices, and they can likely be improved. As an example, both the linear functions and the random MLPs are defined over the space of raw observations, which makes them sensitive to parameterization. Additionally, FRE attempts to encode the entire reward function into a single encoding, whereas state-dependent encodings may be sufficient. The formulation presented in this work requires offline data. Adapting FRE to the online setting will likely require stabilizing the non-stationarity of $q(w|\cdot)$ during training, which we leave for future work.
%%SL.1.29: Here is a potential rephrasing you could incorporate: While FRE provides for flexible and nonlinear policy inference for new tasks, it relies on a hand-specified prior reward distribution. Our specific reward distribution is relatively simple, consists of a mixture of various random functions. However, this choice is somewhat ad hoc, and while we empirically observe that it works well for many benchmark tasks, it remains unclear if there is an optimal and general choice of reward priors....

Generalist agents pretrained in an unsupervised way can enable rapid acquisition of diverse tasks, and FRE provides a step toward training such generalist agents from unsupervised, non-expert offline data. We hope that this will lead to a range of exciting future work that could address acquisition of skill spaces from more diverse data sources (e.g., in robotics, with videos, etc.), further improve the generalization ability of such agents to even more varied downstream tasks, as well as works that provide a deeper theoretical understanding of reward priors and guarantees on downstream task performance, all without the need to hand-specify reward functions or task descriptions during pretraining.
%A useful agent is a generalist that can accomplish a broad ranges of tasks. FRE opens the path to scalable methods of learning generalist agents from \textit{unsupervised, non-expert offline data} by 1) learning a powerful latent reward space, and 2) optimizing policies over this space. There remains fruitful work in investigating the scaling of such techniques to broad domains such as robotics or video data, and discovering the nuances will leave us with powerful methods for generalist agent learning, \textit{without} the need for hand-specified reward functions or task descriptions.

\section*{Impact Statement}
This paper presents work whose goal is to advance the field of Machine Learning. There are many potential societal consequences of our work, none of which we feel must be specifically highlighted here.

\bibliography{example_paper}
\bibliographystyle{icml2024}

%%%%%%%%%%%%%%%%%%%%%%%%%%%%%%%%%%%%%%%%%%%%%%%%%%%%%%%%%%%%%%%%%%%%%%%%%%%%%%%
%%%%%%%%%%%%%%%%%%%%%%%%%%%%%%%%%%%%%%%%%%%%%%%%%%%%%%%%%%%%%%%%%%%%%%%%%%%%%%%
% APPENDIX
%%%%%%%%%%%%%%%%%%%%%%%%%%%%%%%%%%%%%%%%%%%%%%%%%%%%%%%%%%%%%%%%%%%%%%%%%%%%%%%
%%%%%%%%%%%%%%%%%%%%%%%%%%%%%%%%%%%%%%%%%%%%%%%%%%%%%%%%%%%%%%%%%%%%%%%%%%%%%%%
\newpage
\appendix
\onecolumn

\section{Hyperparameters}

\begin{table}[H]
  \centering
  \begin{tabular}{l|l}
    \toprule
    Batch Size & 512 \\
    Encoder Training Steps & 150,000 (1M for ExORL/Kitchen) \\
    Policy Training Steps & 850,000 (1M for ExORL/Kitchen) \\
    Reward Pairs to Encode & 32 \\
    Reward Pairs to Decode & 8 \\
    \midrule
    Ratio of Goal-Reaching Rewards & 0.33 \\
    Ratio of Linear Rewards & 0.33 \\
    Ratio of Randomm MLP Rewards & 0.33 \\
    \midrule
    Number of Reward Embeddings & 32 \\
    Reward Embedding Dim & 128 \\
    \midrule
    Optimizer & Adam \\
    Learning Rate & 0.0001 \\
    RL Network Layers & [512, 512, 512] \\
    Decoder Network Layers & [512, 512, 512] \\
    Encoder Layers & [256, 256, 256, 256] \\
    Encoder Attention Heads & 4 \\
    $\beta$ KL Weight & 0.01 \\
    Target Update Rate & 0.001 \\
    Discount Factor & 0.88 \\
    AWR Temperature & 3.0 \\
    IQL Expectile & 0.8 \\
    \bottomrule
  \end{tabular}
  \vspace{0.1cm}

\caption{Hyperparameters used for FRE.}
\label{table:hyperparameters}
\end{table}

\section{Training Details}

Random goal-reaching functions are generated by sampling random goals from the offline dataset. Specifically we utilize a hindsight experience relabelling \cite{andrychowicz2017hindsight} distribution in accordance with \cite{park2023hiql}. Given a random selected state, we utilize this state as the goal with a $0.2$ chance, a future state within the trajectory with a $0.5$ chance, and a completely random state with a $0.3$ chance. Reward is set to -1 for every timestep that the goal is not achieved. A \textit{done} mask is set to True when the goal is achieved. We ensure that at least one of the samples contains the goal state during the encoding process.

Random Linear functions are generated according to a uniform vector within -1 and 1. On AntMaze, we remove the XY positions from this generation as the scale of the dimensions led to instability. A random binary mask is applied with a 0.9 chance to zero the vector at that dimension, to encourage sparsity and bias towards simpler functions.

Random MLP functions are generated using a neural network of size (state\_dim, 32, 1). Parameters are sampled using a normal distribution scaled by the average dimension of the layer. A tanh activation is used between the two layers. The final output of the neural network is clipped between -1 and 1.

\newpage
\section{Environment Details}

\subsection{AntMaze}
We utilize the \textit{antmaze-large-diverse-v2} dataset from D4RL \cite{fu2020d4rl}. Online evaluation is performed with a length of 2000 timesteps. The ant robot is placed in the center of the maze to allow for more diverse behavior, in comparison to the original start position in the bottom-left. 

For the goal-reaching tasks, we utilize a reward function that considers the goal reached if an agent reaches within a distance of 2 with the target position. The FRE, GC-IQL, GC-BC, and OPAL agents all utilize a discretized preprocessing procedure, where the X and Y coordinates are discretized into 32 bins. 

\subsection{ExORL} 

We utilize \textit{cheetah-run, cheetah-walk, cheetah-run-backwards, cheetah-walk-backwards} and \textit{walker-run, walker-walk} as evaluation tasks. Agents are evaluated for 1000 timesteps. For goal-reaching tasks, we select five consistent goal states from the offline dataset.

FRE assumes that reward functions must be pure functions of the environment state. Because the Cheetah and Walker environments utilize rewards that are a function of the underlying \textit{physics}, we append information about the physics onto the offline dataset during encoder training. Specifically, we append the values of
\begin{verbatim}
self.physics.horizontal_velocity()
self.physics.torso_upright()
self.physics.torso_height()
\end{verbatim} to Walker, and
\begin{verbatim}
self.physics.speed()
\end{verbatim} to Cheetah.

The above auxiliary information is neccessary only for the encoder network, in order to define the true reward functions of the ExORL tasks, which are based on physics states. We found that performance was not greatly affected whether or not the value functions and policy networks have access to the auxilliary information, and are instead trained on the underlying observation space of the environment.

Goals in ExORL are computed when the Euclidean distance between the current state and the goal state is less than 0.1. Each state dimension is normalized according to the standard deviation along that dimension within the offline dataset. Augmented information is not utilized when calculating goal distance.

\subsection{Kitchen}
For the Kitchen evaluation tasks, we utilize the seven standard subtasks within the D4RL Kitchen environment. Because each task already defines a sparse reward, we directly use those sparse rewards as evaluation tasks.

\section{Extended Results}

\begin{table}[H]
  \centering
  \begin{tabular}{l|lllllll}
    \toprule
    % \cmidrule(r){1-2}
    Eval Task & FRE-all & FRE-goals & FRE-lin & FRE-mlp & FRE-lin-mlp & FRE-goal-mlp & FRE goal-lin \\
    \midrule
goal-reaching & \boldmath{$48.8 \pm 6$} & \boldmath{$66.0 \pm 4$} & $6.0 \pm 1$ & $24.0 \pm 6$ & $8.0 \pm 4$ & \boldmath{$52.0 \pm 6$} & \boldmath{$54.0 \pm 12$} \\
directional & \boldmath{$55.2 \pm 8$} & $6.6 \pm 13$ & \boldmath{$55.5 \pm 6$} & $-6.6 \pm 14$ & $47.9 \pm 6$ & $5.1 \pm 25$ & \boldmath{$67.1 \pm 5$} \\
random-simplex & \boldmath{$21.3 \pm 4$} & \boldmath{$23.5 \pm 6$} & $14.4 \pm 3$ & \boldmath{$18.5 \pm 6$} & $14.8 \pm 4$ & \boldmath{$19.7 \pm 5$} & $10.7 \pm 3$ \\
path-all & \boldmath{$63.8 \pm 10$} & $8.3 \pm 11$ & $50.5 \pm 9$ & \boldmath{$65.4 \pm 5$} & $58.5 \pm 7$ & $58.6 \pm 23$ & $55.8 \pm 8$ \\
\midrule
total & \boldmath{$47.3 \pm 7$} & $26.1 \pm 8$ & $31.6 \pm 5$ & $25.3 \pm 8$ & $32.3 \pm 5$ & $33.8 \pm 15$ & $46.9 \pm 7$ \\
    \bottomrule
  \end{tabular}
  \vspace{0.1cm}

\caption{Full results comparing FRE agents trained on different subsets of random reward functions in AntMaze. }
\label{table:transfer}
\end{table}

\begin{figure}[H]
    \centering
    \includegraphics[width=0.8\textwidth]{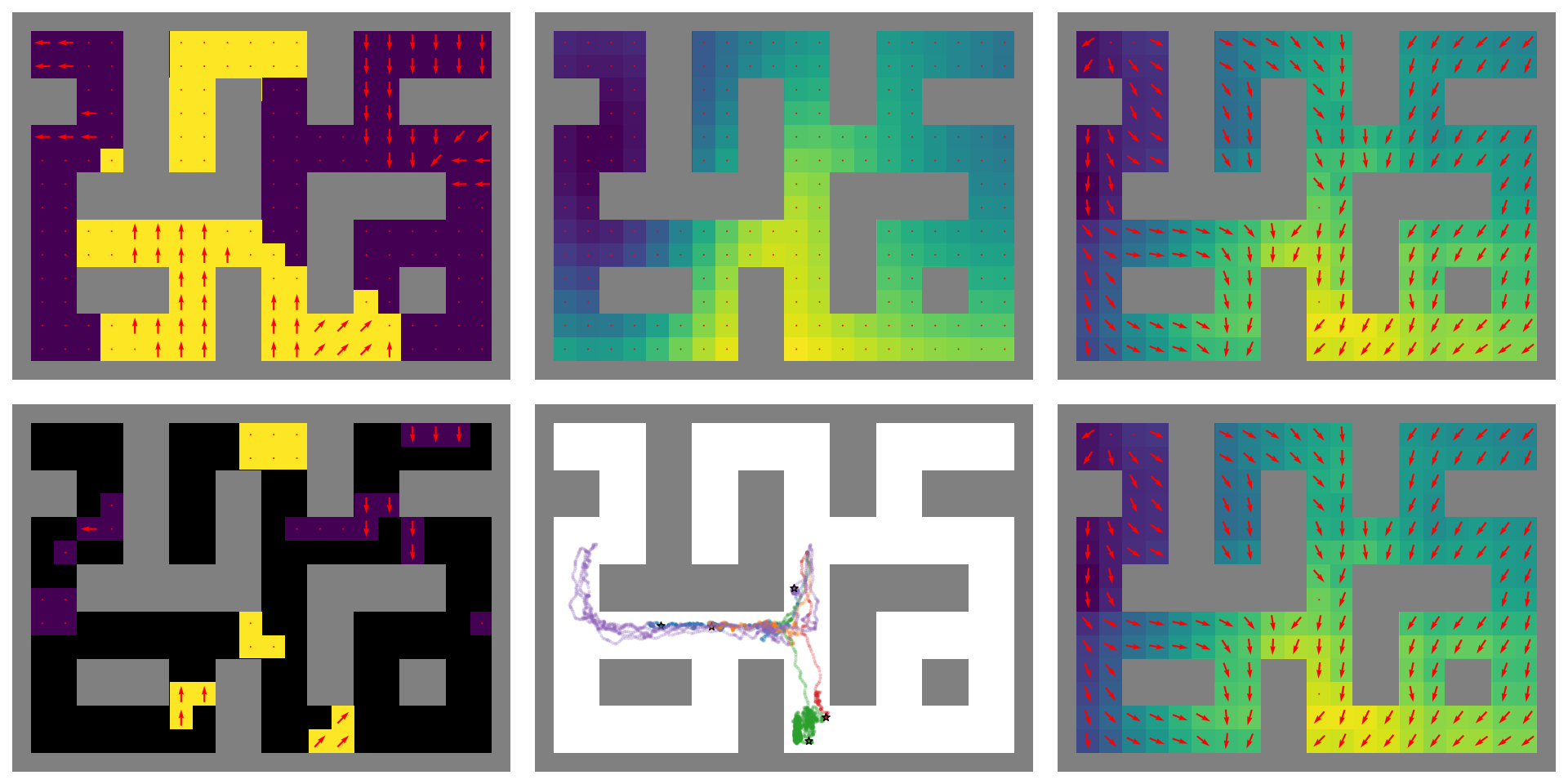}
    \includegraphics[width=0.8\textwidth]{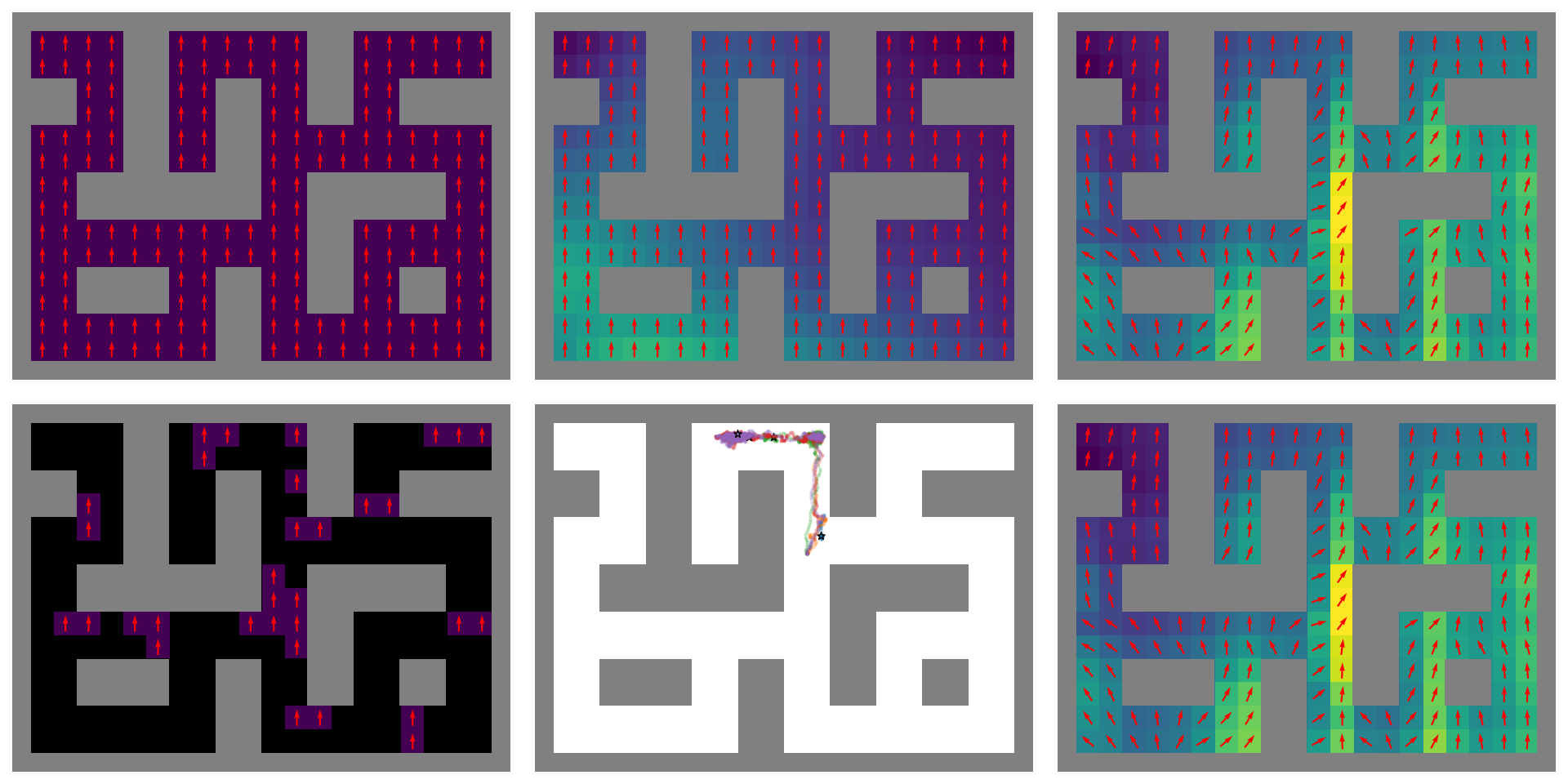}
    \includegraphics[width=0.8\textwidth]{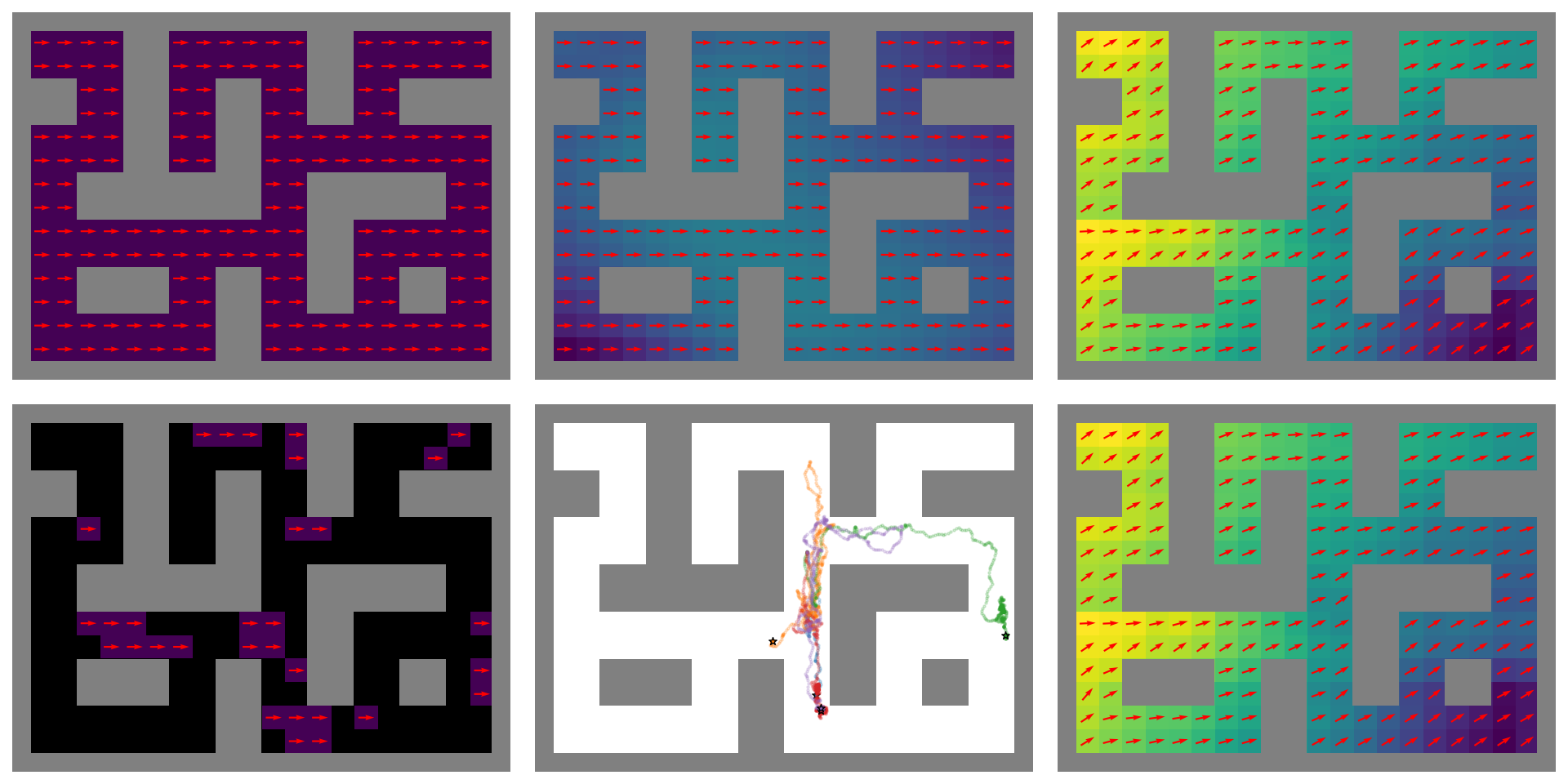}

    \caption{Additional examples of FRE results on AntMaze. Arranged three examples per page. For each run, from top-left to bottom-right: True reward function, predicted reward, Q function 1, randomly sampled states for encoding, policy trajectory, Q function 2.}
\end{figure}

\begin{figure}[H]
    \centering
    \includegraphics[width=0.8\textwidth]{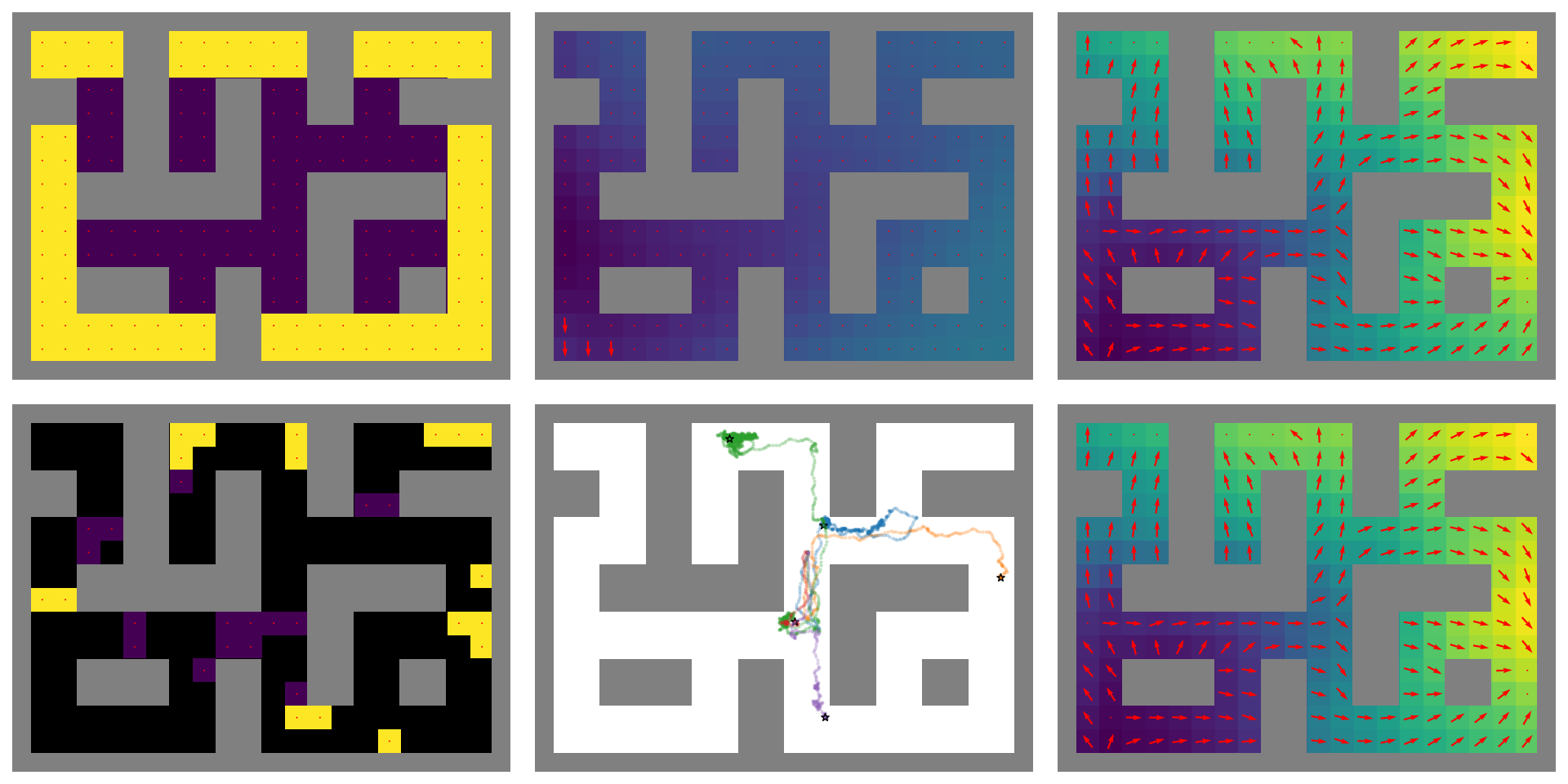}
    \includegraphics[width=0.8\textwidth]{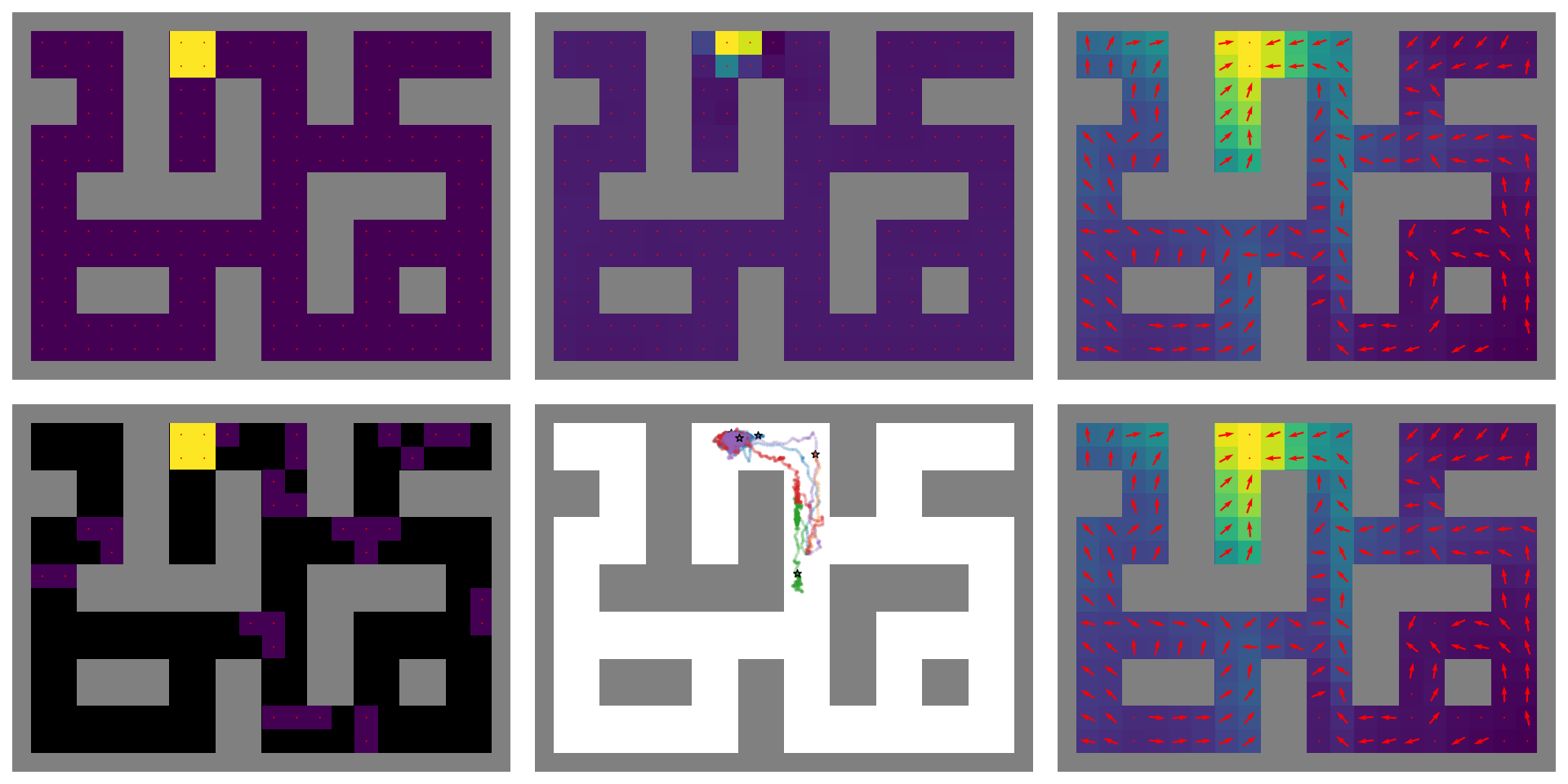}
    \includegraphics[width=0.8\textwidth]{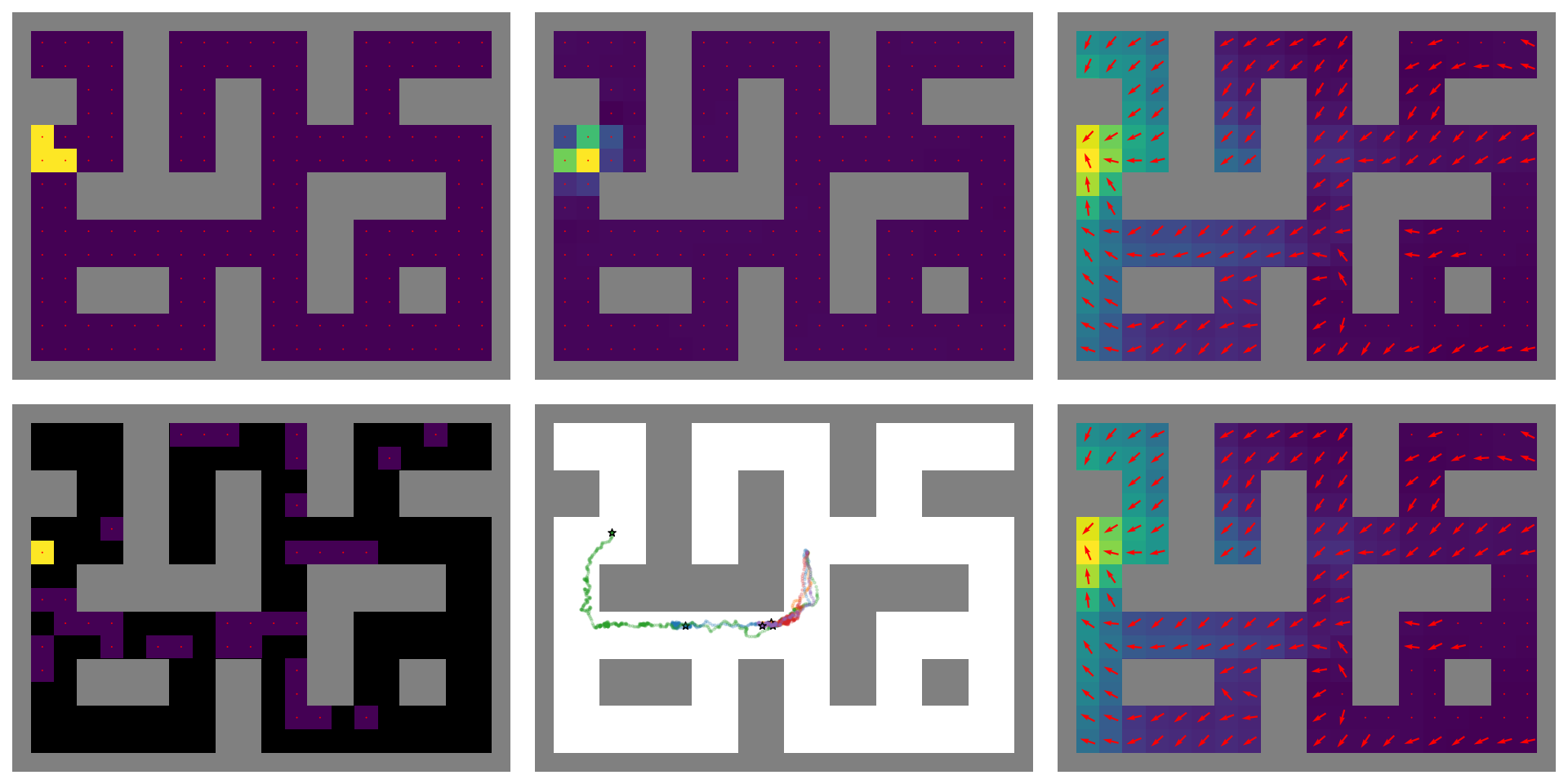}

    \caption{Additional examples of FRE results on AntMaze. Arranged three examples per page. For each run, from top-left to bottom-right: True reward function, predicted reward, Q function 1, randomly sampled states for encoding, policy trajectory, Q function 2.}
\end{figure}

\begin{figure}[H]
    \centering
    \includegraphics[width=0.8\textwidth]{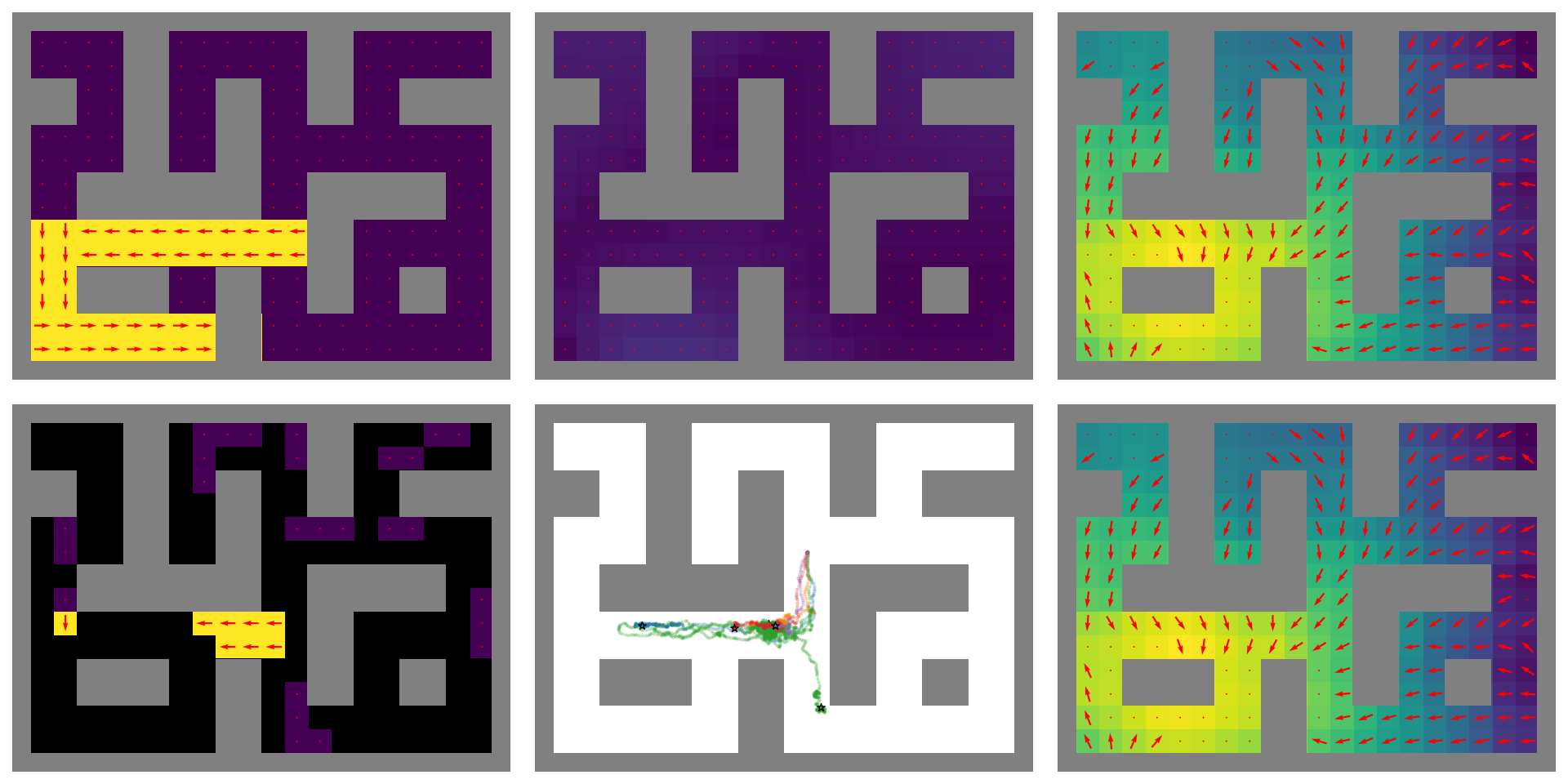}
    \includegraphics[width=0.8\textwidth]{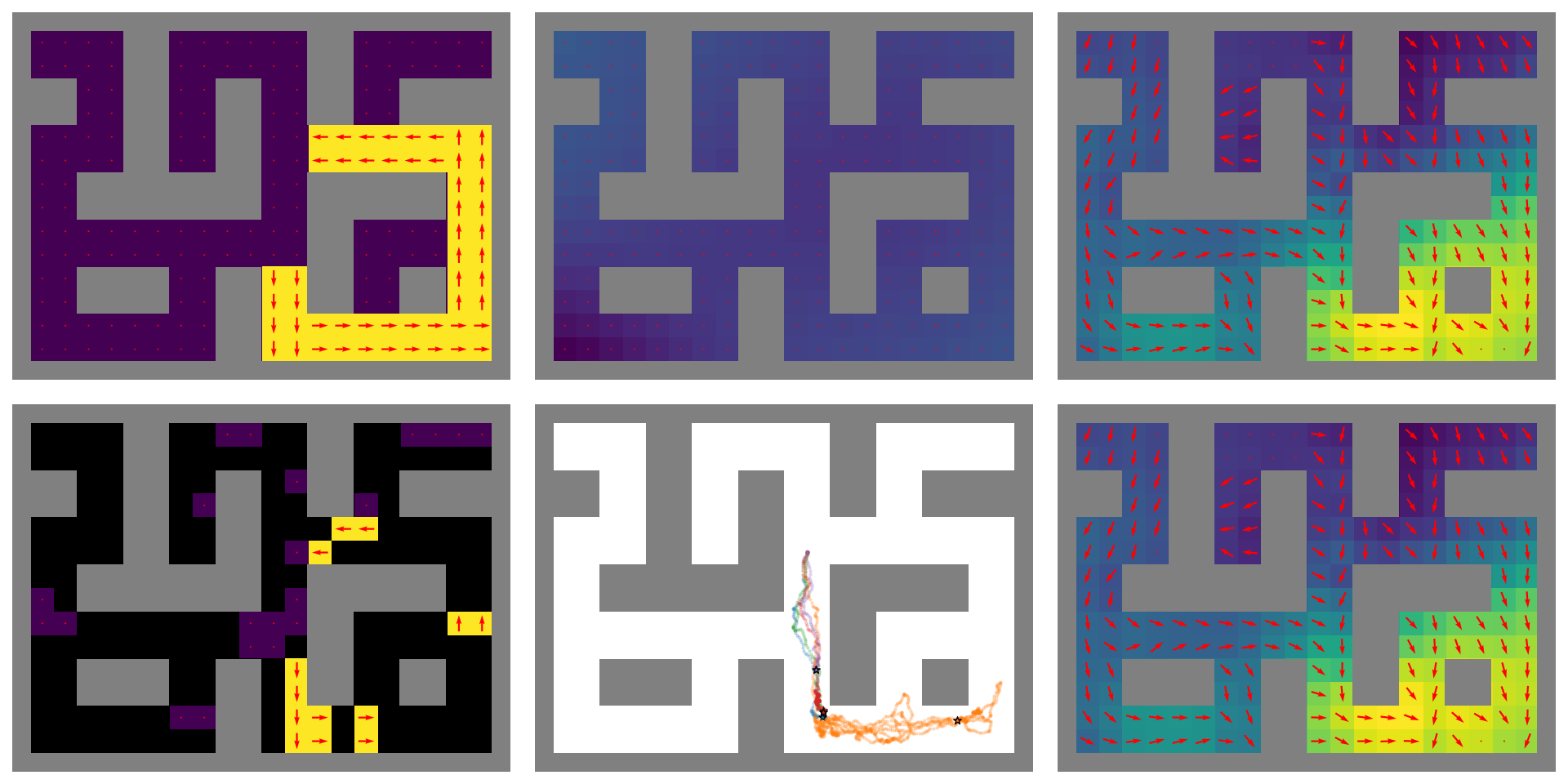}
    \includegraphics[width=0.8\textwidth]{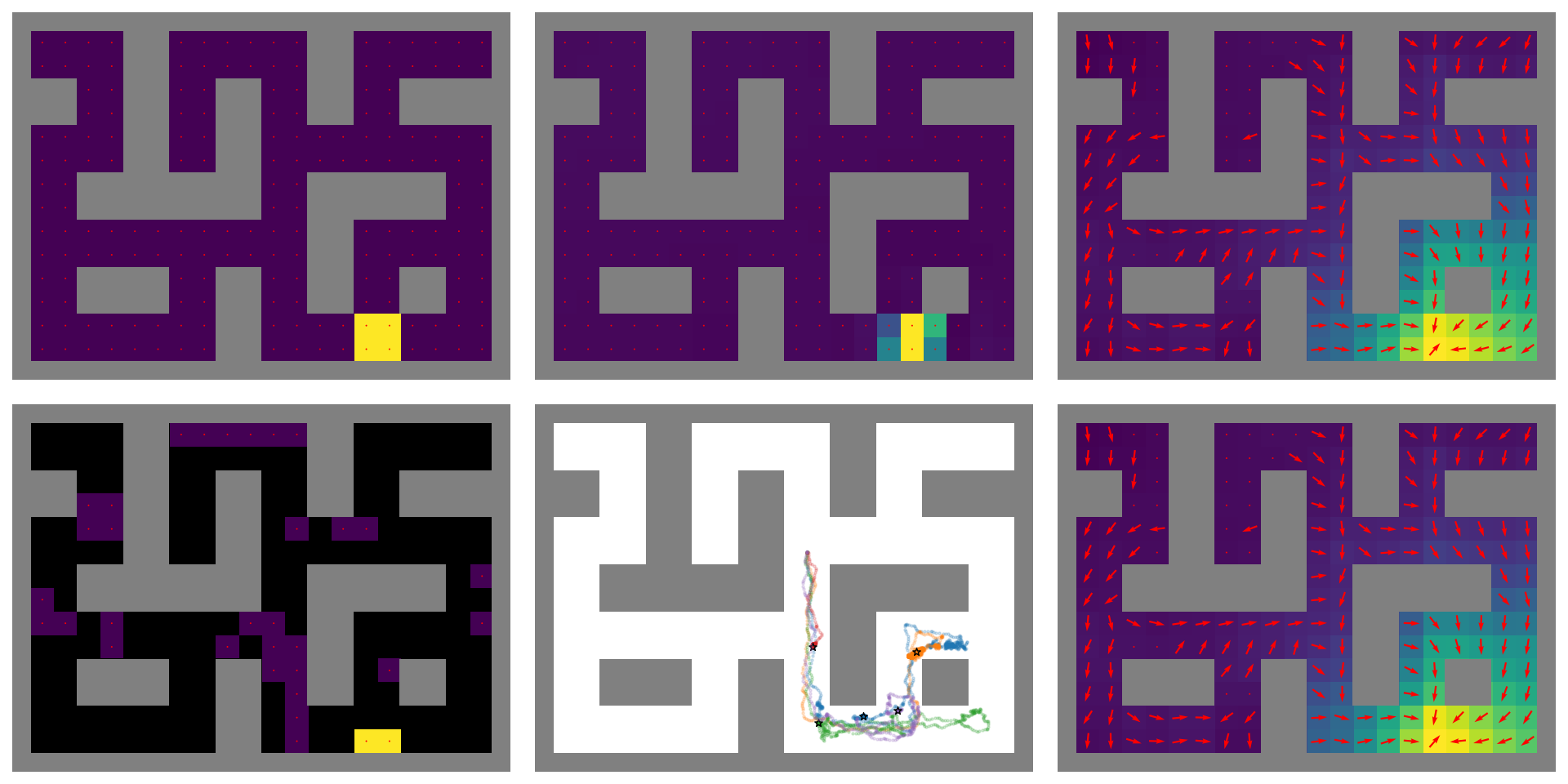}

    \caption{Additional examples of FRE results on AntMaze. Arranged three examples per page. For each run, from top-left to bottom-right: True reward function, predicted reward, Q function 1, randomly sampled states for encoding, policy trajectory, Q function 2.}
\end{figure}

\end{document}